\documentclass[pdflatex,sn-mathphys-num]{sn-jnl}%
\usepackage{graphicx}%
\usepackage{multirow}%
\usepackage{amsmath,amssymb,amsfonts}%
\usepackage{amsthm}%
\usepackage{mathrsfs}%
\usepackage[title]{appendix}%
\usepackage[dvipsnames]{xcolor} % more colors
\usepackage{textcomp}%
\usepackage{manyfoot}%
\usepackage{booktabs}%
\usepackage{algorithm}%
\usepackage{listings}%
\usepackage{xspace}
\newcommand{\oos}{\textsc{oos}\xspace}
\newcommand{\ood}{\textsc{ood}\xspace}
\usepackage{subfig}
\usepackage[version=4]{mhchem}
\usepackage{algorithmic}

\theoremstyle{thmstyleone}%
\theoremstyle{thmstyletwo}%
\theoremstyle{thmstylethree}%

\begin{document}

\title[Article Title]{Known Unknowns: Out-of-Distribution Property Prediction in Materials and Molecules}

\author*[1]{\fnm{Nofit} \sur{Segal}}
\email{pulkitag@mit.edu}
%\email{nofit@mit.edu}

\author*[2]{\fnm{Aviv} \sur{Netanyahu}}
\email{rafagb@mit.edu}
%\email{avivn@mit.edu}

\author[3, 4]{\fnm{Kevin P.} \sur{Greenman}}
%\email{kgreenman@catholic.tech}

\author[2]{\fnm{Pulkit} \sur{Agrawal}}
% \email{pulkitag@mit.edu}
\equalcont{Equal Advising.}

\author[1]{\fnm{Rafael} \sur{Gómez-Bombarelli}}
% \email{rafagb@mit.edu}
\equalcont{Equal Advising.}

\affil[1]{\orgdiv{Materials Science and Engineering}, \orgname{MIT}, \orgaddress{\street{Memorial Dr}, \city{Cambridge}, \postcode{02139}, \state{MA}, \country{USA}}}

\affil[2]{\orgdiv{Electrical Engineering and Computer Science}, \orgname{MIT}, \orgaddress{\street{Vassar St}, \city{Cambridge}, \postcode{02139}, \state{MA}, \country{USA}}}

\affil[3]{\orgdiv{Chemical Engineering}, \orgname{Catholic Institute of Technology}, \orgaddress{\street{Broadway}, \city{Cambridge}, \postcode{02142}, \state{MA}, \country{USA}}}

\affil[4]{\orgdiv{Chemistry}, \orgname{Catholic Institute of Technology}, \orgaddress{\street{Broadway}, \city{Cambridge}, \postcode{02142}, \state{MA}, \country{USA}}}

\abstract{Discovery of high-performance materials and molecules requires identifying extremes with property values that fall outside the known distribution. Therefore, the ability to extrapolate to out-of-distribution (\ood) property values is critical for both solid-state materials and molecular design. Our objective is to train predictor models that extrapolate zero-shot to higher ranges than in the training data, given the chemical compositions of solids or molecular graphs and their property values. We propose using a transductive approach to \ood property prediction, achieving improvements in prediction accuracy. In particular, the True Positive Rate (TPR) of \ood classification of materials and molecules improved by 3x and 2.5x, respectively, and precision improved by 2x and 1.5x compared to non-transductive baselines. Our method leverages analogical input-target relations in the training and test sets, enabling generalization beyond the training target support, and can be applied to any other material and molecular tasks.}

\keywords{machine learning, materials property prediction, extrapolation, out-of-distribution, transduction}

\maketitle
\section{Introduction}
\label{sec1}

Designing new materials and molecules is essential for the development of new technologies. Traditionally, this design process involves extensive experimental iteration or high-throughput methods to screen databases, which are time-consuming and resource-intensive \cite{axelrod2022learning, sanchez2018inverse}. As a result, there is increasing interest in applying machine learning (ML) techniques to accelerate the discovery of materials and molecules with desired properties \cite{axelrod2022learning, sanchez2018inverse, bilodeau2022generative, noh2020machine, kim2021deep}.
There is particular interest in property values that are outside the known property value distribution as these will most likely lead to discovering new materials that will, in turn, unlock new capabilities and technologies.

One strategy for finding materials and molecules with desired properties is inverse design through conditional generation where materials with out-of-distribution (\ood) property values are unavailable, and the goal is to generate them.\cite{noh2020machine, kim2021deep, zeni2023mattergen, yang2023scalable, xie2021crystal, sanchez2018inverse}.
A complementary approach is screening a large database of candidates based on their predicted properties. \cite{walters2020applications, dunn2020benchmarking, wang2021compositionally, zhuo2018predicting, de2021materials, ward2016general}. In this setting, the objective is to identify \ood materials and molecules from a set of known candidates with unknown property values. 
However, both approaches typically struggle when property values fall outside the training distribution. \cite{noh2020machine,wang2021compositionally,de2021materials, kauwe2020can, zhao2022limitations, omee2024structure}. Enhancing extrapolative capabilities in property prediction would improve the screening of large candidate spaces in terms of precision by identifying promising compounds and molecules with exceptional properties. This approach could help guide further synthesis and computational efforts, ultimately advancing materials and molecular design.

Extrapolation in materials science can refer to both the domain (materials space) and range (property values) of the predictive function. It is often used to refer \textit{generalization} to unseen classes of materials structures and chemical spaces — for example, training on metals and predicting ceramics or training on artificial molecules and predicting natural products. Here, we address extrapolation in material property values.

When \ood generalization is defined with respect to the materials space, extrapolation often reduces to interpolation. This occurs because test sets tend to remain within the same distribution as the training data representation space \cite{li2024probing}. This includes predictive models employing leave-one-cluster-out extrapolation strategies \cite{meredig2018can, muckley2023interpretable, omee2024structure}, as well as generative approaches designed to achieve \ood generalization to structures with different atomic compositions or larger numbers of atoms \cite{yang2024mattersim, merchant2023scaling, batatia2023foundation}.

When \ood generalization is defined with respect to the range of the predictive function, classical machine learning models face significant challenges in extrapolating property predictions through regression. As a result, several approaches have shifted toward classifying \ood materials instead \cite{kauwe2020can, zhao2022limitations}.

\begin{figure*}[htbp]
    \centering
    \includegraphics[width=.99\linewidth]{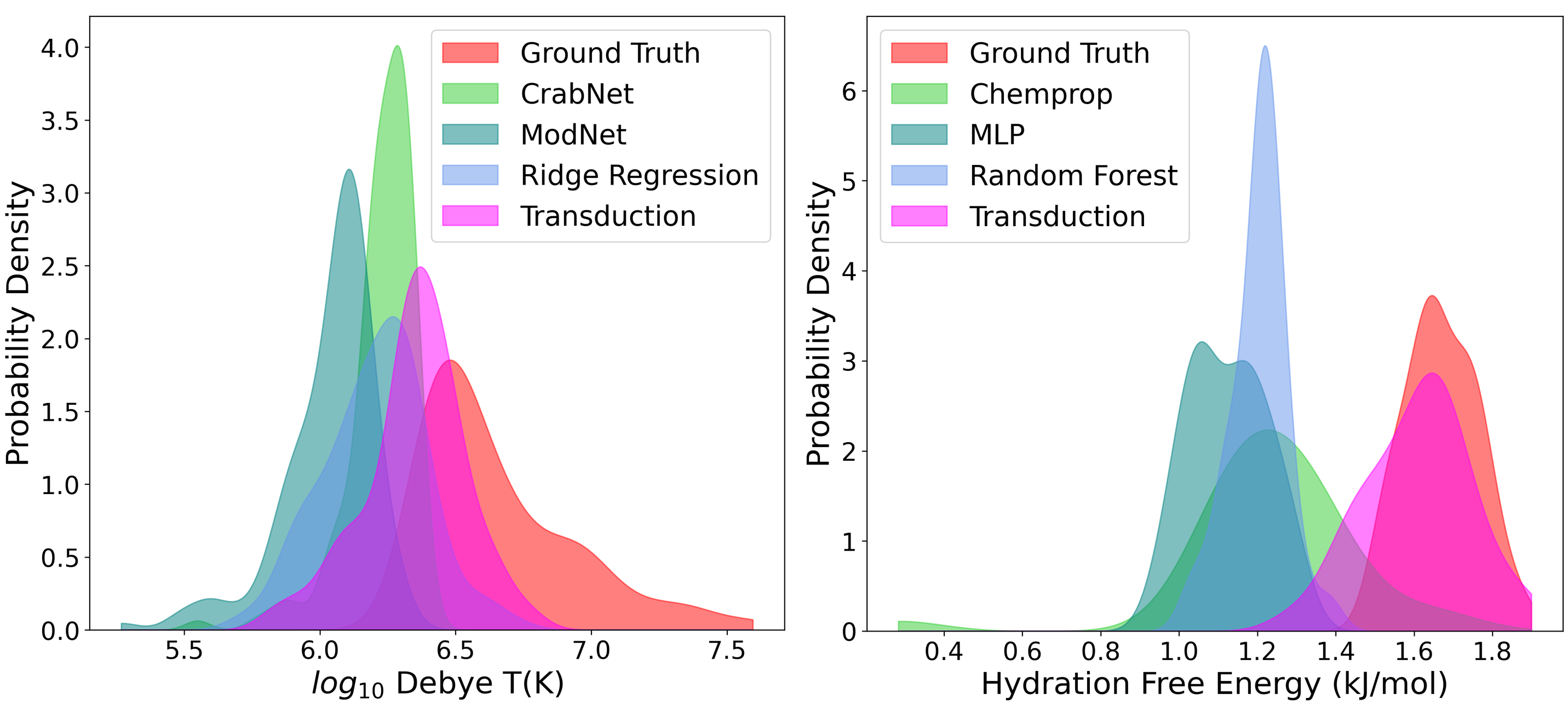}%
    \caption{\textbf{Bilinear Transduction prediction distribution on out-of-distribution values is closer to the ground truth distribution compared with other machine learning methods.}
    (\textbf{Left}) probability density functions for out-of-distribution \textbf{materials} Debye temperature. Compared with CrabNet, ModNet, and Ridge Regression, Bilinear Transduction (purple) has the most overlap with the ground truth (red). 
    (\textbf{Right}) probability density functions for out-of-distribution \textbf{molecular} hydration free energy. Ground truth (red) and machine learning method predictions estimated with kernel density estimation -- Chemprop, MLP, Random Forest, and Bilinear Transduction (purple) which has the most overlap with the ground truth.}
    \label{fig:ood-kde-plot}
\end{figure*}

In this work, we focus on predicting property values from materials representations and exploring zero-shot extrapolation to property value ranges beyond the training distribution.
We adapt the Bilinear Transduction method \cite{netanyahu2023transduction}, which has shown success in extrapolation tasks in other domains, and investigate its effectiveness for property value extrapolation in materials science (Figure~\ref{fig:ood-kde-plot}). 
The core idea of Bilinear Transduction is reparameterizing the prediction problem. Rather than making property value predictions on a new candidate material, predictions are made based on a known training example and the difference in representation space between the two materials. 
During inference, property values are predicted similarly -- based on a chosen training example and the difference between it and the new sample.
This method has the potential to extrapolate by learning \textit{how} property values change as a function of material differences rather than predicting these values from new materials.

Our work advances the field by enhancing extrapolation capabilities, enabling more accurate predictions of material and molecular properties beyond known value ranges. Specifically, our method significantly improves \ood precision, ensuring a higher percentage of high-potential candidates with desirable properties during the screening of large databases. This approach has the potential to streamline the identification process by reducing time and resource expenditure on low-potential candidates, thereby accelerating the discovery of materials and molecules with high synthesis viability.

We present a comprehensive evaluation of our approach using common benchmarks, spanning high-throughput computational and experimental datasets for solid-state materials and molecules.

\section{Results}\label{sec2}

\begin{table*}[htbp]
\small
\begin{center}
\caption{Solids \ood mean average prediction error and standard error of the mean.
}
\label{tab:analysis}
\resizebox{\textwidth}{!}{\begin{tabular}{ccccccc}
\toprule
\textbf{Dataset} & \textbf{Property} & \textbf{\#Samples} &
\textbf{Ridge Reg.} \cite{kauwe2020can} & \textbf{MODNet} \cite{de2021materials} & \textbf{CrabNet} \cite{wang2021compositionally} & \textbf{Ours}\\
\midrule
\multirow{5}{*}{AFLOW} & Band Gap [eV] & 14123 & 2.59 $\pm$ 0.03 & 2.65 $\pm$ 0.04 & \textbf{1.47 $\pm$ 0.03} & 1.51 $\pm$ 0.04\\
\multirow{5}{*}{\cite{curtarolo2012aflow}} & Bulk Modulus [GPa] & 2740 & 74.0 $\pm$ 3.8 & 93.06 $\pm$ 3.7 & 59.25 $\pm$ 3.2 & \textbf{47.4 $\pm$ 3.4}\\
& Debye Temperature [K] & 2740 & 0.45 $ 
\pm$ 0.03 & 0.62 $\pm$ 0.03 & 0.38 $\pm$ 0.02 & \textbf{0.31 $\pm$ 0.02}\\
& Shear Modulus [GPa] & 2740 & 0.69 $ 
\pm$ 0.03 & 0.78 $\pm$ 0.04 & 0.55 $\pm$ 0.02 & \textbf{0.42 $\pm$ 0.02}\\
& Thermal Conductivity [\(\frac{W}{mK}\)] & 2734 & 1.07 $ 
\pm$ 0.05 & 1.5 $\pm$ 0.05 & 0.97 $\pm$ 0.03 & \textbf{0.83 $\pm$ 0.04}\\
& Thermal Expansion [\(K^{-1}\)] & 2733 & 0.44 $ 
\pm$ 0.02 & 0.47 $\pm$ 0.02 & \textbf{0.37 $\pm$ 0.02} & 0.39 $\pm$ 0.02\\
\midrule
\multirow{2}{*}{Matbench} & Band Gap [eV] & 2154 & 6.37 $\pm$ 0.28 & 3.26 $\pm$ 0.13 & 2.70 $\pm$ 0.13 & \textbf{2.54 $\pm$ 0.16}\\
\multirow{2}{*}{\cite{dunn2020benchmarking}} & Refractive Index & 4764 & 14.4 $\pm$ 2.0 & 4.24 $\pm$ 0.48 & 3.92 $\pm$ 0.5 & \textbf{3.81 $\pm$ 0.49}\\
& Yield Strength [MPa] & 312 & 972 $\pm$ 34 & 731 $\pm$ 82 & 740 $\pm$ 49 & \textbf{591 $\pm$ 62}\\
\midrule
\multirow{2}{*}{MP} & Bulk Modulus [GPa] & 6307 & 151 $\pm$ 14 & 60.1 $\pm$ 3.9 & 57.8 $\pm$ 4.2 & \textbf{45.8 $\pm$ 3.9} \\
\multirow{2}{*}{\cite{jain2013commentary}} & Elastic Anisotropy & 6331 & 165 $\pm$ 17 & 60.0 $\pm$ 4.5 & 61.4 $\pm$ 4.6 & \textbf{59.8 $\pm$ 4.5} \\
& Shear Modulus [GPa] & 6184 & 134.5 $\pm$ 7.2 & 65.6 $\pm$ 2.5 & 65.3 $\pm$ 2.8 & \textbf{63.2 $\pm$ 2.6}\\
\bottomrule
\end{tabular}}
\end{center}
\end{table*}

\begin{table*}[htbp]
\small
\begin{center}
\caption{Solids 30\% extrapolative precision
}
\label{tab:30prec}
\resizebox{\textwidth}{!}{\begin{tabular}{ccccccc}
\toprule
\textbf{Dataset} & \textbf{Property} & \textbf{Ridge Reg.} \cite{kauwe2020can}& \textbf{MODNet} \cite{de2021materials} & \textbf{CrabNet} \cite{wang2021compositionally}& \textbf{Ours}\\
\midrule
\multirow{5}{*}{AFLOW} & Band Gap & 0.16 & 0.15 & 0.14 & \textbf{0.22} \\
\multirow{5}{*}{\cite{curtarolo2012aflow}}
 & Bulk Modulus & 0.22 & 0.30 & 0.17 & \textbf{0.40} \\
 & Debye Temperature & 0.19 & 0.06 & 0.07 & \textbf{0.20} \\
 & Shear Modulus & \textbf{0.07} & \textbf{0.07} & 0.06 & \textbf{0.07} \\
 & Thermal Conductivity & \textbf{0.15} & 0.05 & 0.06 & 0.09 \\
 & Thermal Expansion & 0.26 & 0.28 & 0.15 & \textbf{0.36} \\
\midrule
\multirow{2}{*}{Matbench} & Band Gap & 0.05 & 0.11 & 0.17 & \textbf{0.20} \\
\multirow{2}{*}{\cite{dunn2020benchmarking}}
 & Refractive Index & 0.17 & 0.32 & 0.20 & \textbf{0.40} \\
 & Yield Strength & 0.00 & 0.00 & 0.00 & \textbf{0.67} \\
\midrule
\multirow{2}{*}{MP} & Bulk Modulus & 0.22 & 0.21 & 0.14 & \textbf{0.60} \\
\multirow{2}{*}{\cite{jain2013commentary}}
 & Elastic Anisotropy & 0.01 & \textbf{0.09} & 0.03 & 0.03 \\
 & Shear Modulus & 0.28 & 0.27 & 0.20 & \textbf{0.53} \\
\bottomrule
\end{tabular}}
\end{center}
\end{table*}

\subsection{Solids}
\label{subsec:mat}

We demonstrate the extrapolation capability of bilinear transduction on three common benchmarks for solid materials property prediction tasks involving 12 prediction tasks, with dataset sizes ranging between $\sim$300 and $\sim$14000, and compared against three baselines. 

The solids datasets include material compositions and their property values.
\textbf{AFLOW} contains material property values obtained from high-throughput calculations \cite{curtarolo2012aflow}. Following \citet{kauwe2020can}, who evaluated classical ML algorithms on AFLOW, we curate a subset of six properties: band gap, bulk modulus, Debye temperature, shear modulus, thermal conductivity, and thermal expansion, out of which the last four are scaled by applying a base 10 logarithm.
\textbf{Matbench} is an automated leaderboard for benchmarking ML algorithms predicting solid material properties \cite{dunn2020benchmarking}. Matbench contains three composition-based regression tasks: experimentally measured band gap \cite{zhuo2018predicting}, experimentally measured yield strength of steels \cite{CiteDrive2022}, and calculated refractive index \cite{petousis2017high}. 
\textbf{Materials Project (MP)} provides materials and their property values derived from high-throughput calculations \cite{jain2013commentary}.
Following \citet{wang2021compositionally}, we focus on bulk modulus, shear modulus, and ratio of elastic anisotropy. In cases of duplicate compositions, we retain the entry with the target value corresponding to the lowest formation enthalpy.

We compare with \textbf{Ridge Regression}, the strongest method in \citet{kauwe2020can}, who evaluate classical ML algorithms on \ood property values, and also \textbf{MODNet} \cite{de2021materials} and \textbf{CrabNet} \cite{wang2021compositionally}, all of which use stoichiometry as input and are leading models in composition-based property prediction.

Table~\ref{tab:analysis} presents the mean average error (MAE) for \ood predictions on solids and molecules. Bilinear transduction consistently outperforms or performs comparably to the baseline methods. 

In addition, we report model performance on a purely extrapolative task of discovering high-performance extremes:  identifying the 30\% of test samples with the highest property values, which we refer to as \textit{the top \ood candidates} (Table~\ref{tab:30prec}). Specifically, it measures the fraction of true top \ood candidates correctly identified among the top predicted \ood candidates. The score is calculated as the ratio of true top \ood candidates included in the top predictions to the total number of predicted top candidates. The held-out set is composed of both in-distribution and \ood samples in equal parts (see Section~\ref{methods}), so the 30\% of the test dataset corresponds to 60\% of the \ood-sourced part of the test set. Furthermore, this metric re-weights misranked in-distribution errors 19-fold, reflecting the 95:5 split between in-distribution validation and \ood test sets.

\begin{figure*}[htbp]
    \centering
    % First figure
    \includegraphics[width=\linewidth]{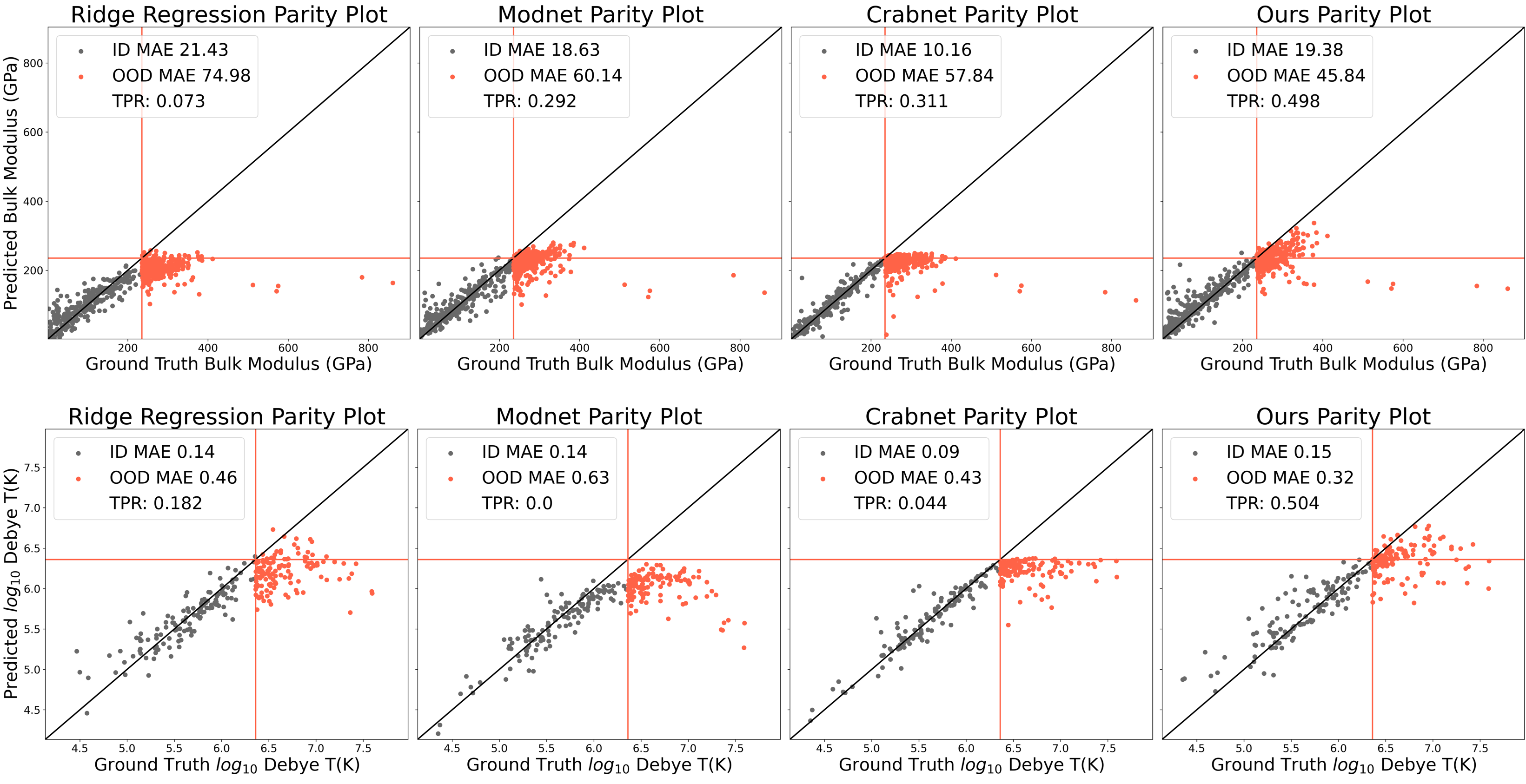}
    % Vertical spacing between figures
    \caption{\textbf{In-distribution and out-of-distribution Bulk Modulus (top) and Debye Temperature (bottom) predictions vs. ground truth values}. 
    While Ridge Regression \cite{kauwe2020can}, MODNet \cite{de2021materials}, CrabNet \cite{wang2021compositionally}, and Bilinear Transduction (ours), perform well within the training distribution (gray dots bounded by the red horizontal line), Bilinear Transduction extends predictions beyond this range on \ood data (red dots) closest to the ground truth, achieving a lower \ood MAE and higher TPR.}
    \label{fig:comparison-plots-solids}
\end{figure*}

Figure~\ref{fig:ood-kde-plot} demonstrates that Bilinear Transduction produces a prediction for band gap distribution that is closer to the \ood ground truth distribution.
Figure~\ref{fig:comparison-plots-solids} shows that Bilinear Transduction extrapolates to some extent, whereas the other baselines do not exceed the training support threshold.

\subsection{Molecules}
\label{subsec:mol}
For molecules, we sourced one dataset involving 4 graph-to-property prediction tasks with dataset sizes ranging from $\sim$600 to 4200 and compared against three baselines. 
\vspace{-2mm}

\begin{table*}[htbp]
\small
\begin{center}
\caption{Molecules \ood mean average prediction error and standard error of the mean.
}
\label{tab:mol_analysis}
\resizebox{\textwidth}{!}{\begin{tabular}{ccccccc}
\toprule
\textbf{Dataset} & \textbf{Property} & \textbf{\#Samples} & \textbf{Chemprop} \cite{heid2023chemprop} & \textbf{Random Forest} \cite{breiman2001random} & \textbf{MLP} \cite{gardner1998artificial} & \textbf{Ours}\\
\midrule
\multirow{3}{*}{MoleculeNet} & ESOL [\(\frac{mol}{L}\)] & 1128 & 0.47 $\pm$ 0.04 & 0.67 $\pm$ 0.04 & 0.5 $\pm$ 0.04 & \textbf{0.42 $\pm$ 0.04}\\
\multirow{3}{*}{\cite{Ramsundar-et-al-2019}} & Freesolv [\(\frac{kJ}{mol}\)] & 643 & 0.44 $\pm$ 0.03 & 0.42 $\pm$ 0.02 & 0.5 $\pm$ 0.02 & \textbf{0.08 $\pm$ 0.01}\\
& Lipophilicity [$\log D$] & 4200 & 0.75 $ \pm$ 0.02 & 1.02 $\pm$ 0.02 & 0.9 $\pm$ 0.02 & \textbf{0.7 $\pm$ 0.02}\\
& BACE binding [IC50] & 1513 & 1.03 $ \pm$ 0.06 & 0.93 $\pm$ 0.05 & 0.95 $\pm$ 0.07 & \textbf{0.73 $\pm$ 0.05}\\
\bottomrule
\end{tabular}}
\end{center}
\end{table*}

\begin{table*}[htbp]
\small
\begin{center}
\caption{Molecules 30\% extrapolative precision
}
\label{tab:30prec_mol}
\resizebox{\textwidth}{!}{\begin{tabular}{ccccccc}
\toprule
\textbf{Dataset} & \textbf{Property} & \textbf{Chemprop} \cite{heid2023chemprop} & \textbf{Random Forest} \cite{breiman2001random} & \textbf{MLP} \cite{gardner1998artificial} & \textbf{Ours}\\
\midrule
\multirow{3}{*}{MoleculeNet} 
 & ESOL & \textbf{0.17} & 0.11 & 0.15 & 0.16 \\
 \multirow{3}{*}{\cite{Ramsundar-et-al-2019}}
 & Freesolv & 0.50 & 0.22 & 0.44 & \textbf{0.83} \\
 & Lipophilicity & 0.05 & 0.05 & \textbf{0.07} & \textbf{0.07} \\
 & BACE binding & 0.04 & \textbf{0.07} & 0.05 & 0.06 \\
\bottomrule
\end{tabular}}
\end{center}
\end{table*}

\textbf{MoleculeNet} includes molecular graphs encoded as SMILES representations \cite{weininger1988smiles} and their property values derived from high-throughput calculations and experimental trials \cite{Ramsundar-et-al-2019}. We focus on physical chemistry and biophysics properties suitable for regression -- ESOL, Freesolv, Lipophilicity and BACE binding affinity.

We compare with \textbf{Random Forest (RF)} \cite{breiman2001random}, a classical ML tree-based method, and \textbf{Multi Layer Perceptron (MLP)} which use RDKit descriptors as input \cite{rdkit}. These serve as ablations of our method, using the same representation with partial structural information as we do. We also compare with \textbf{Chemprop} \cite{heid2023chemprop}, a leading method for property prediction from molecular graphs via message-passing. Chemprop learns representations from structural information that is not explicitly available in the descriptor-based representation.

\begin{figure*}[htbp]
    \centering
    \includegraphics[width=\linewidth]{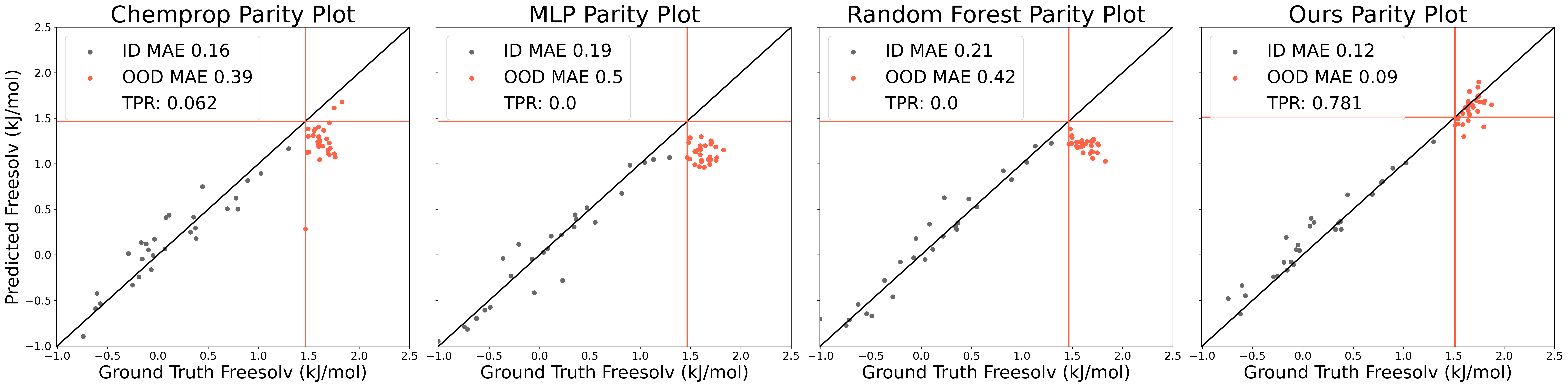}
    \caption{\textbf{In-distribution, and out-of-distribution Freesolv predictions vs. ground truth values}. 
    While Chemprop \cite{heid2023chemprop}, RF, MLP and Bilinear Transduction (ours), perform well within the training distribution (gray dots bounded by the red horizontal line), only Bilinear Transduction performs well beyond this range on \ood data (red dots).}
    \label{fig:comparison-plots-mol}
\end{figure*}

Table~\ref{tab:mol_analysis} presents the mean average error (MAE) for \ood predictions on solids and molecules. Bilinear transduction consistently outperforms or performs comparably to the baseline methods. 
Figure~\ref{fig:ood-kde-plot} demonstrates that Bilinear Transduction produces a prediction for the Freesolv data distribution that is closer to the \ood ground truth distribution and Figure~\ref{fig:comparison-plots-mol} shows that Bilinear Transduction extrapolates, whereas the other baselines rarely surpass the boundary marking the beginning of the test range. The fraction of true top \ood candidates correctly identified among the top predicted \ood candidates is presented in Table~\ref{tab:30prec_mol}.

\subsection{Learning Using Analogies}
\label{subsec:analogies}
The success of transduction in these tasks is related to one of the tenets of chemistry: \textit{similar materials have similar properties}. The transductive approach effectively builds on this idea, by proposing that \textit{similar changes in chemical compounds or molecular structure} imply \textit{similar changes in properties}. At inference, our approach selects the anchor by minimizing the difference between $\Delta x_{\mathrm{te,an}}$ and $\Delta x_{\mathrm{tr}}$. We demonstrate how these algebraic operations in the embedding space relate to chemical changes as measured in the domain. 

\begin{figure*}[h!]
    \centering
    % Large PCA plot on the left
    \includegraphics[width=0.99\linewidth]{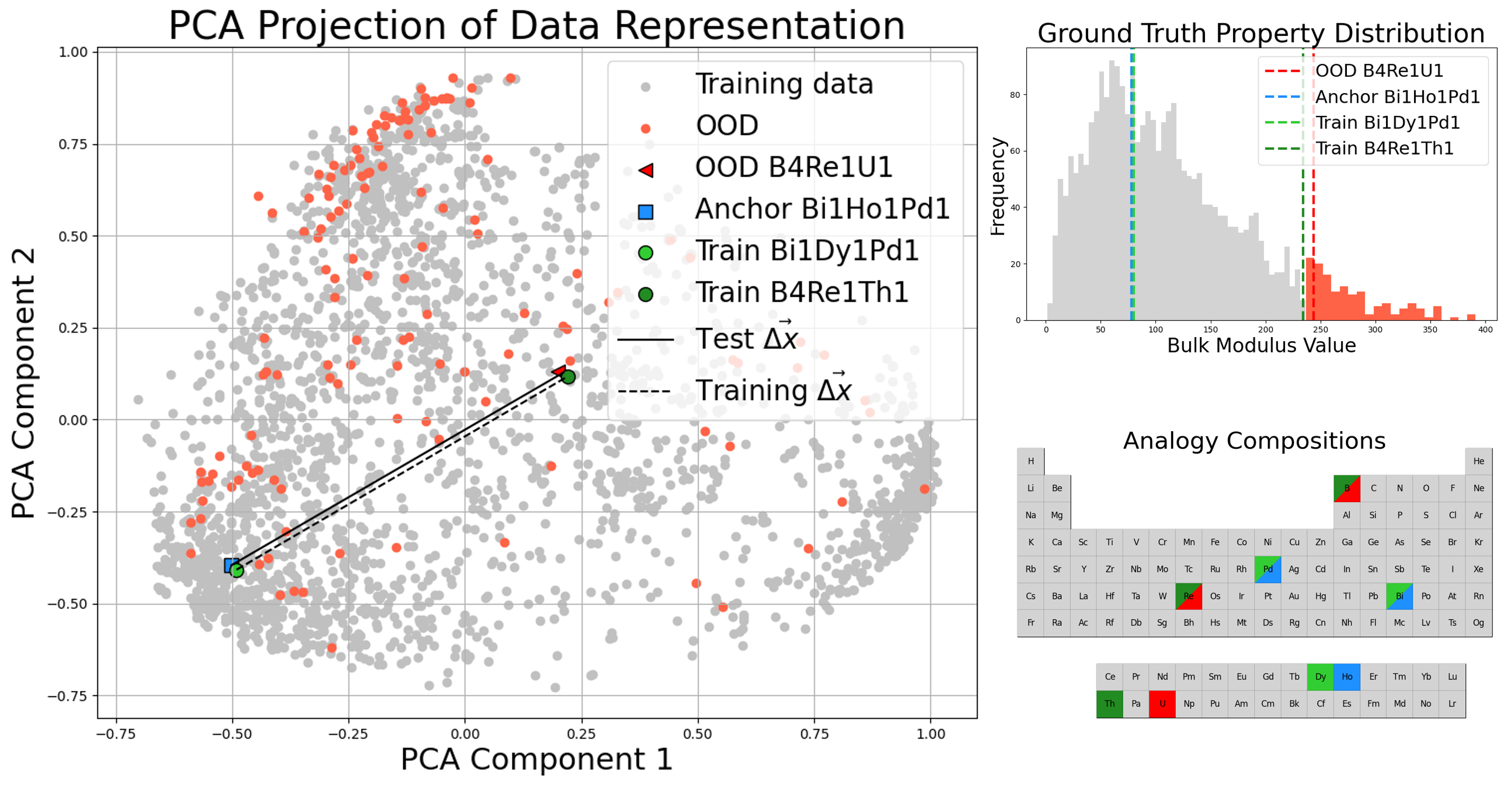}
    \caption{\textbf{Analogy Visualization Solids}. AFLOW bulk modulus \ood predictions are based on in-distribution anchors, that paired with \ood points, form analogies to training pairs.
    (a) PCA plot of all samples in the dataset. \textcolor{red}{\ood}-\textcolor{Cerulean}{anchor} difference is similar to \textcolor{ForestGreen}{training}-\textcolor{LimeGreen}{anchor} difference.
    (b) Ground truth bulk modulus \textcolor{gray}{training} and \textcolor{RedOrange}{test} distributions and \textcolor{red}{\ood}, \textcolor{Cerulean}{anchor}, and analogous \textcolor{ForestGreen}{training} point and \textcolor{LimeGreen}{anchor} values.
    (c) Analogy compositional visualization. \textcolor{red}{\ood} and \textcolor{ForestGreen}{training} points differ by one neighboring f-block element. So do \textcolor{Cerulean}{\ood} and \textcolor{LimeGreen}{training} anchors.
    }
    \label{fig:analogies0}
\end{figure*}

For solids, these operations are expressed as elemental changes. Figure~\ref{fig:analogies0} demonstrates this in bulk modulus inference of an \ood sample with stoichiometry \ce{B_{4}ReU}. The model selects in-distribution \ce{BiHoPd} as the anchor, analogous to training anchor \ce{BiDyPd} and training target \ce{B_{4}ReTh}. Specifically, the compositions of these anchors (\ce{BiDyPd} and \ce{BiHoPd}), and the targets (\ce{B_{4}ReTh} and \ce{B_{4}ReU}), differ by only one neighboring f-block element: \ce{Dy (Z=66)} to \ce{Ho (Z=67)}, and \ce{Th (Z=90)} to \ce{U (Z=92)}. See Appendix~\ref{subsubssec:appendix_analogies_solids} for more examples. 

\begin{figure*}[h!]
    \centering
    \includegraphics[width=.99\linewidth]{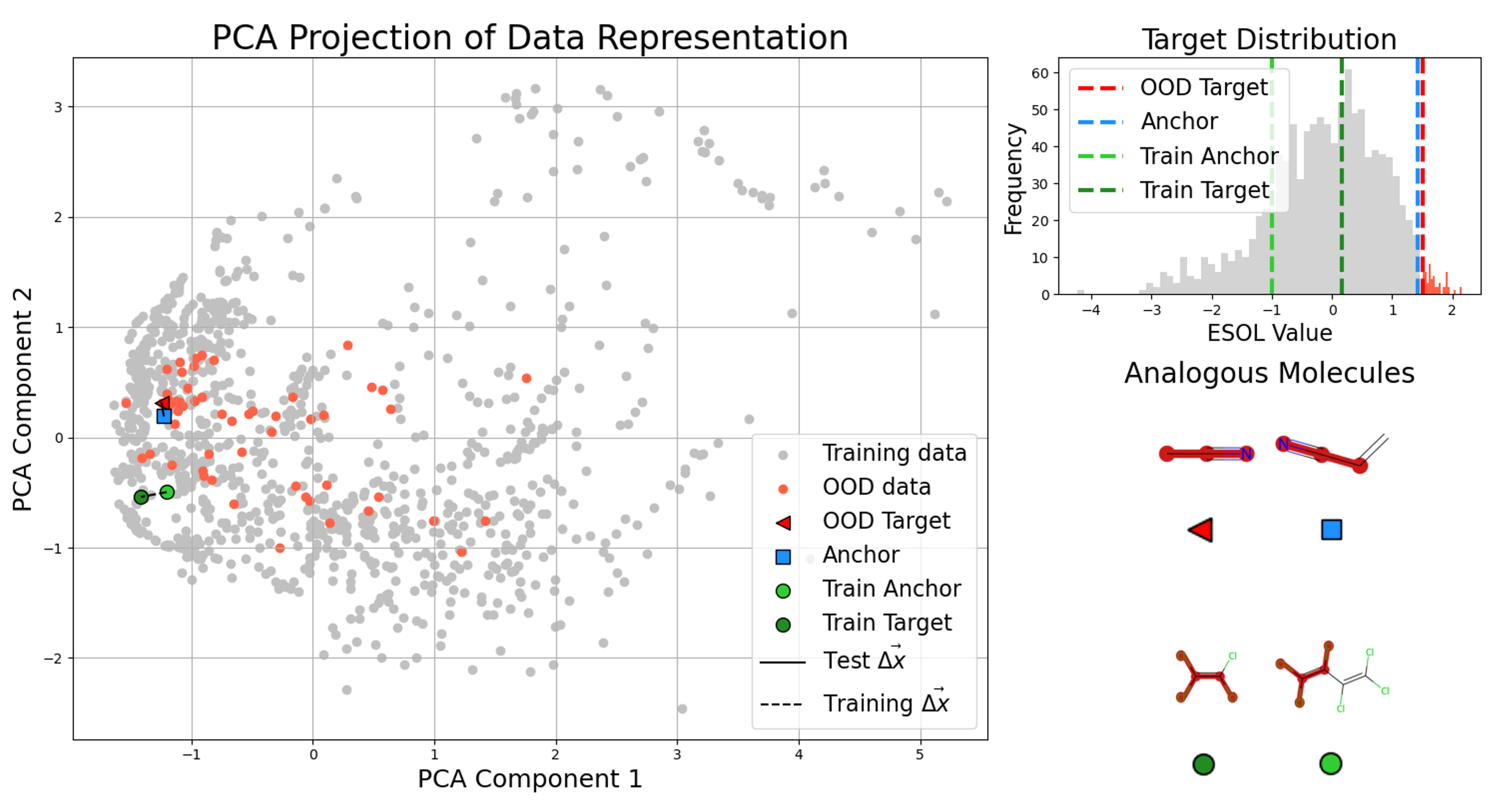}
    \caption{\textbf{Analogy Visualization Molecules}. MoleculeNet ESOL \ood predictions are based on differences between in-distribution anchors and \ood points, that form analogies to training pairs.
    (a) \textcolor{red}{\ood}-\textcolor{Cerulean}{anchor} and training \textcolor{ForestGreen}{target}-\textcolor{LimeGreen}{anchor} differences.
    (b) Ground truth \textcolor{gray}{training} and \textcolor{RedOrange}{test} distributions and \textcolor{red}{\ood}, \textcolor{Cerulean}{anchor}, and analogous training \textcolor{ForestGreen}{target}, \textcolor{LimeGreen}{anchor} values.
    (c) Analogous molecule pairs. \textcolor{red}{\ood}-\textcolor{Cerulean}{anchor} and training \textcolor{ForestGreen}{target}-\textcolor{LimeGreen}{anchor} similarities measured with MCS highlighted in red.
    }
    \label{fig:analogies_mol}
\end{figure*}

For molecules, these operations manifest as structural similarity. This is notable given that the RDKit descriptor vector used as input lacks detailed structural and connectivity information from the SMILES representation. 
Figure~\ref{fig:analogies_mol} illustrates this in ESOL inference, with the Maximum Common Structure (MCS) highlighted (Figure~\ref{fig:analogies_mol}\textcolor{blue}{c}) between the anchor-\ood pair, and the training anchor-target pair. Each pair’s structures show high similarity, differing by the addition of a conjugated double bond that extends the molecular backbone. See Appendix~\ref{subsubssec:appendix_analogies_mol} for additional examples.

\newpage
% [htbp]

\newpage
\section{Methods}\label{methods}

Let $\mathcal{X}\subseteq\mathbb{R}^n$ denote a representation space of data points, such as RDKit descriptors derived from Simplified Molecular Input Line Entry System (SMILES) for molecules, and composition-based descriptors derived from elemental properties for solids. We choose fixed descriptor-based representations as they offer interpretable features, useful for smaller datasets that are common in materials science. Additionally, for solids, a composition-based approach enables more robust predictions, as it implies weaker assumptions about the material \cite{damewood2023representations}.
Let $\mathcal{Y}\subseteq\mathbb{R}$ denote the space of labels -- in our case, material property values. 
We split the data into training, validation and test sets such that \ood test samples $\mathcal{D}_{\mathcal{Y}}^{\mathrm{te}}\subseteq\mathcal{Y}$ are $5\%$ of the data with the highest property values, an in-distribution validation set is a random sample of $5\%$ of the remaining data, and the training set $\mathcal{D}^{\mathrm{tr}}=\{(x,y)\}_n\subseteq\mathcal{X}\times\mathcal{Y}$ is the rest.

We adapt Bilinear Transduction \cite{netanyahu2023transduction}, a regression method, which offers the following formulation (see description in Appendix~\ref{appendix-blt} and adapted version summary in Algorithm~\ref{alg:blt-matex}).
Let $\Delta \mathcal{X}=\{x_i-x_j|x_i,x_j\in\mathcal{X}\}$ denote the differences distribution. During training, rather than predicting $h_{\theta}:\mathcal{X}\to\mathcal{Y}$, predictor $h_{\theta}:\Delta \mathcal{X}\times\mathcal{X}\to\mathcal{Y}$ predicts value $y_2\in\mathcal{D}^{\mathrm{tr}}_\mathcal{Y}$ for data point $x_2\in\mathcal{D}^{\mathrm{tr}}_\mathcal{X}$ given anchor point $x_1\in\mathcal{D}^{\mathrm{tr}}_\mathcal{X}$ and the difference between them $\Delta x_{21}=x_2-x_1\in\mathcal{D}^{\mathrm{tr}}_{\Delta \mathcal{X}}$. 
The predictor $h_{\theta}(\Delta x,x)=f_{\theta}(\Delta x)g_{\theta}(x)$ is implemented as a bilinear function in non-linear embeddings of $\Delta x$ and $x$. 
I.e. during training we predict the property value $y_i$ of material $x_i$ from material $x_j$ and their difference $x_i-x_j$, where material $x_j$ has a lower property value.

Given test point $x_{\mathrm{te}}\in\mathcal{D}_{\mathcal{X}}^{\mathrm{te}}$, its predicted value is $h_{\theta}(\Delta x_{\mathrm{te,an}},x_{\mathrm{an}})$. Anchor $x_{\mathrm{an}}\in\mathcal{D}^{\mathrm{tr}}_\mathcal{X}$ is the training point that minimizes the distance between its difference with the test point $\Delta x_{\mathrm{te,an}}=x_{\mathrm{te}}-x_{\mathrm{an}}$, and differences within the training distribution $\mathcal{D}^{\mathrm{tr}}_{\Delta X}$. Formally, $x_{\mathrm{an}}=\mathrm{argmin}_{x_i\in\mathcal{D}^{\mathrm{tr}}}\{||\Delta x_{\mathrm{te},i}-\Delta x||_2\mid \Delta x\in\mathcal{D}^{\mathrm{tr}}_{\Delta X}\}$.

Descriptor-based features encapsulate fundamental chemical and physical information that directly influences materials and molecular characteristics. Therefore, the difference between feature vectors is related, possibly intricately, to the change in property value.
Bilinear Transduction has the potential to extrapolate by learning \textit{how} property values change as a function of compositional differences instead of predicting these values directly.

\begin{algorithm}[!h]
    \begin{algorithmic}[1]
    \STATE{}\textbf{Input:} Training set $(x_1,y_1),\dots,(x_n,y_n)$
    \STATE{}\textbf{Train:} Train $\theta$ on loss $
    \mathcal{L}(\theta) = \textstyle\sum_{i=1}^n\sum_{j:y_j<y_i} \ell(h_{\theta}(x_i -x_j,x_j),y_i)$
    \STATE{}\textbf{Test:} For each new $x_{\mathrm{te}}$, let $x_{\mathrm{an}} = \mathrm{argmin}_{x_{\mathrm{an}}\in\mathcal{D}^{\mathrm{tr}}_\mathcal{X}}\{\|x_{\mathrm{te}} - x_{\mathrm{an}} - \Delta x_{\mathrm{tr}}\|_2 \,\mid\, \Delta x_{\mathrm{tr}} \in \mathcal{D}^{\mathrm{tr}}_{\Delta X}\}$, and predict 
    \begin{align*}
    &y = h_{\theta}(x_{\mathrm{te}} - x_{\mathrm{an}},x_{\mathrm{an}})        
    \end{align*}
    \end{algorithmic}
    \caption{Bilinear Transduction for Molecules and Materials} 
    \label{alg:blt-matex}
\end{algorithm}

\section{Data availability}
The raw data, along with processing scripts, are available at \url{https://github.com/learningmatter-mit/matex}.

\section{Code availability}
An open-source implementation of Matex is available at \url{https://github.com/learningmatter-mit/matex}.

\backmatter

\clearpage

\section{Supplementary Information}

\subsection{Additional Results}
\label{appendix-add}
\subsubsection{In-Distribution Predictions}
In Table~\ref{tab:analysis-in-dist} we demonstrate that all methods perform well, making Bilinear Transduction a strong predictor both in and out of distribution alike.

\begin{table*}[htbp]
\small
\begin{center}
\caption{
Solids (top) and molecules (bottom) \textbf{In-distribution} mean average prediction error and standard error of the mean.
}
\label{tab:analysis-in-dist}
\resizebox{\textwidth}{!}{\begin{tabular}{cccccc}
\toprule
\textbf{Dataset} & \textbf{Property} &\textbf{Ridge Reg.} \cite{kauwe2020can} & \textbf{MODNet} \cite{de2021materials} & \textbf{CrabNet} \cite{wang2021compositionally} & \textbf{Ours}\\
\midrule
\multirow{5}{*}{AFLOW} & Band Gap [eV] &  0.87 $\pm$ 0.04 & 0.56 $\pm$ 0.02 & 0.35 $\pm$ 0.02 & 0.61 $\pm$ 0.02\\
\multirow{5}{*}{\cite{curtarolo2012aflow}} & Bulk Modulus [GPa] & 15.41 $\pm$ 1.21 & 15.1 $\pm$ 1.3 &  8.01 $\pm$ 1.05 & 13.06 $\pm$ 1.6\\
& Debye Temperature [K] & 0.13 $\pm$ 0.01 & 0.13 $\pm$ 0.01 & 0.09 $\pm$ 0.01 & 0.14 $\pm$ 0.01\\
& Shear Modulus [GPa] & 0.31 $\pm$ 0.03 & 0.27 $\pm$ 0.02 & 0.19 $\pm$ 0.02 & 0.31 $\pm$ 0.03\\
& Thermal Conductivity [\(\frac{W}{mK}\)] & 0.47 $\pm$ 0.03 & 0.4 $\pm$ 0.02 & 0.31 $\pm$ 0.03 & 0.43 $\pm$ 0.04\\
& Thermal Expansion [\(K^{-1}\)] & 0.11 $\pm$ 0.01 & 0.07 $\pm$ 0.01 & 0.04 $\pm$ 0.0 & 0.11 $\pm$ 0.01\\
\midrule
\multirow{2}{*}{Matbench} & Band Gap [eV] & 1.75 $\pm$ 0.07 & 0.32 $\pm$ 0.03 & 0.24 $\pm$ 0.03 & 0.49 $\pm$ 0.05 \\
\multirow{2}{*}{\cite{dunn2020benchmarking}} & Refractive Index & 1.00 $\pm$ 0.05 & 0.15 $\pm$ 0.01 & 0.13 $\pm$ 0.02 & 0.16 $\pm$ 0.01 \\
& Yield Strength [MPa] & 411 $\pm$ 75 & 62.5 $\pm$ 11.8 & 52.4 $\pm$ 18.1 & 156 $\pm$ 33\\
\midrule
\multirow{2}{*}{MP} & Bulk Modulus [GPa] & 36.9 $\pm$ 1.21 & 18.63 $\pm$ 1.16 & 10.2 $\pm$ 0.8 & 19.4 $\pm$ 1.3\\
\multirow{2}{*}{\cite{jain2013commentary}} & Elastic Anisotropy & 22.00 $\pm$ 2.01 & 2.12 $\pm$ 0.34 & 1.24 $\pm$ 0.06 & 2.4 $\pm$ 0.3\\
& Shear Modulus [GPa] & 35.7 $\pm$ 1.2 & 12.8 $\pm$ 0.7 & 8.75 $\pm$ 0.63 & 13.6 $\pm$ 0.7\\
\midrule
\midrule
& & \textbf{Chemprop} \cite{heid2023chemprop} & \textbf{Random Forest} \cite{breiman2001random} & \textbf{MLP} \cite{gardner1998artificial} & \\
\midrule
\multirow{3}{*}{MoleculeNet} & ESOL [\(\frac{mol}{L}\)] & 0.28 $\pm$ 0.03 & 0.25 $\pm$ 0.03 & 0.28 $\pm$ 0.03 & 0.29 $\pm$ 0.04\\
\multirow{3}{*}{\cite{Ramsundar-et-al-2019}} & Freesolv [\(\frac{kJ}{mol}\)] & 0.16 $\pm$ 0.02 & 0.20 $\pm$ 0.06 & 0.18 $\pm$ 0.06 & 0.12 $\pm$ 0.02\\
& Lipophilicity [$\log D$] & 0.36 $ \pm$ 0.02 & 0.40 $\pm$ 0.02 & 0.38 $\pm$ 0.03 & 0.46 $\pm$ 0.03\\
& BACE binding [IC50] & 0.45 $ \pm$ 0.04 & 0.37 $\pm$ 0.04 & 0.43 $\pm$ 0.05 & 0.51 $\pm$ 0.05\\
\bottomrule
\end{tabular}}
\end{center}
\end{table*}

\subsubsection{True Positive Rate (TPR)}
In Table~\ref{tab:tpr}, we compare the TPR of \ood detection, showcasing our ability to produce candidate materials with outstanding property values, with an improvement of 3x compared with the strongest baseline.

\begin{table*}[htbp]
\small
\begin{center}
\caption{
Solids (top) and molecules (bottom) \ood \textbf{True Positive Rate} (TPR).
}
\label{tab:tpr}
\resizebox{\textwidth}{!}{\begin{tabular}{cccccc}
\toprule
\textbf{Dataset} & \textbf{Property} &\textbf{Ridge Reg.} \cite{kauwe2020can} & \textbf{MODNet} \cite{de2021materials} & \textbf{CrabNet} \cite{wang2021compositionally} & \textbf{Ours}\\
\midrule
\multirow{5}{*}{AFLOW} & Band Gap [eV] & 0.0 & 0.0 & 0.032 & \textbf{0.132}\\
\multirow{5}{*}{\cite{curtarolo2012aflow}} & Bulk Modulus [GPa] & 0.124 & 0.007 & 0.199 & \textbf{0.336}\\
& Debye Temperature [K] & 0.182 & 0.0 & 0.0 & \textbf{0.504}\\
& Shear Modulus [GPa] & 0.058 & 0.088 & 0.044 & \textbf{0.182}\\
& Thermal Conductivity [\(\frac{W}{mK}\)] & 0.088 & 0.0 & 0.0 & \textbf{0.132}\\
& Thermal Expansion [\(K^{-1}\)] & 0.154 & 0.154 & 0.132 & \textbf{0.191}\\
\midrule
\multirow{2}{*}{Matbench} & Band Gap [eV] & 0.0 & 0.004 & 0.0 & \textbf{0.009}\\
\multirow{2}{*}{\cite{dunn2020benchmarking}} & Refractive Index & 0.009 & 0.101 & 0.004 & \textbf{0.122}\\
& Yield Strength [MPa] & \textbf{0.0} & \textbf{0.0} & \textbf{0.0} & \textbf{0.0}\\
\midrule
\multirow{2}{*}{MP} & Bulk Modulus [GPa] & 0.073 & 0.003 & 0.311 & \textbf{0.498} \\
\multirow{2}{*}{\cite{jain2013commentary}} & Elastic Anisotropy & 0.0 & 0.0 & 0.0 & \textbf{0.006} \\
& Shear Modulus [GPa] & 0.0 & 0.0 & 0.003 & \textbf{0.084}\\
\midrule
\midrule
& & \textbf{Chemprop} \cite{heid2023chemprop} & \textbf{Random Forest} \cite{breiman2001random} & \textbf{MLP} \cite{gardner1998artificial} & \\
\midrule
\multirow{3}{*}{MoleculeNet} & ESOL [\(\frac{mol}{L}\)] & \textbf{0.357} & 0.0 & 0.196 & 0.268\\
\multirow{3}{*}{\cite{Ramsundar-et-al-2019}} & Freesolv [\(\frac{kJ}{mol}\)] & 0.062 & 0.0 & 0.0 & \textbf{0.781} \\
& Lipophilicity [$\log D$] & 0.024 & 0.0 & 0.014 & \textbf{0.057}\\
& BACE binding [IC50] & 0.0 & 0.0 & 0.0 & \textbf{0.013}\\
\bottomrule
\end{tabular}}
\end{center}
\end{table*}

\subsubsection{Analogy Analysis}
\label{appendix-analogies}

\begin{figure*}[htbp]
    \centering
    % Large PCA plot on the left
    \includegraphics[width=0.99\linewidth]{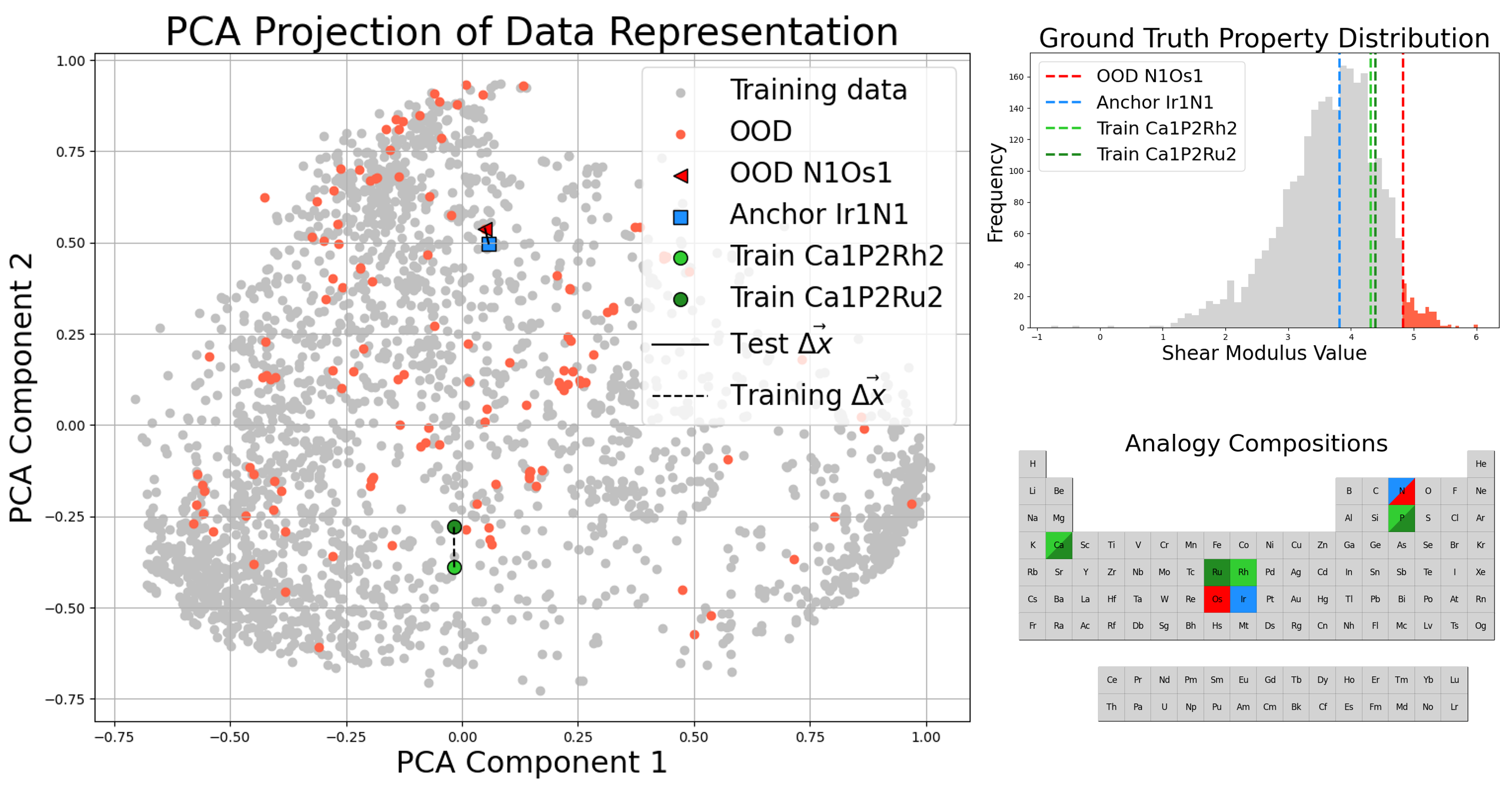}
    \caption{\textbf{Visualizing Analogies in Bilinear Transduction}. AFLOW shear modulus \ood predictions are based on in-distribution anchors, that paired with \ood points, form analogies to training pairs.
    (a) PCA plot of all samples in the dataset. The difference between the \ood point and its anchor is similar to the difference between the training point and anchor. 
    (b) Ground truth shear modulus training (gray) and test (red) distributions and \ood, anchor, and analogous training pair values.
    (c) Analogy compositional visualization. \textcolor{Cerulean}{anchor} and \textcolor{red}{\ood} differ by one neighboring d-block element. So do training \textcolor{LimeGreen}{anchor} and \textcolor{ForestGreen}{target}.
    }
    \label{fig:appendix_analogies1}
\end{figure*}

\begin{figure*}[htbp]
    \centering
    \includegraphics[width=0.98\linewidth]{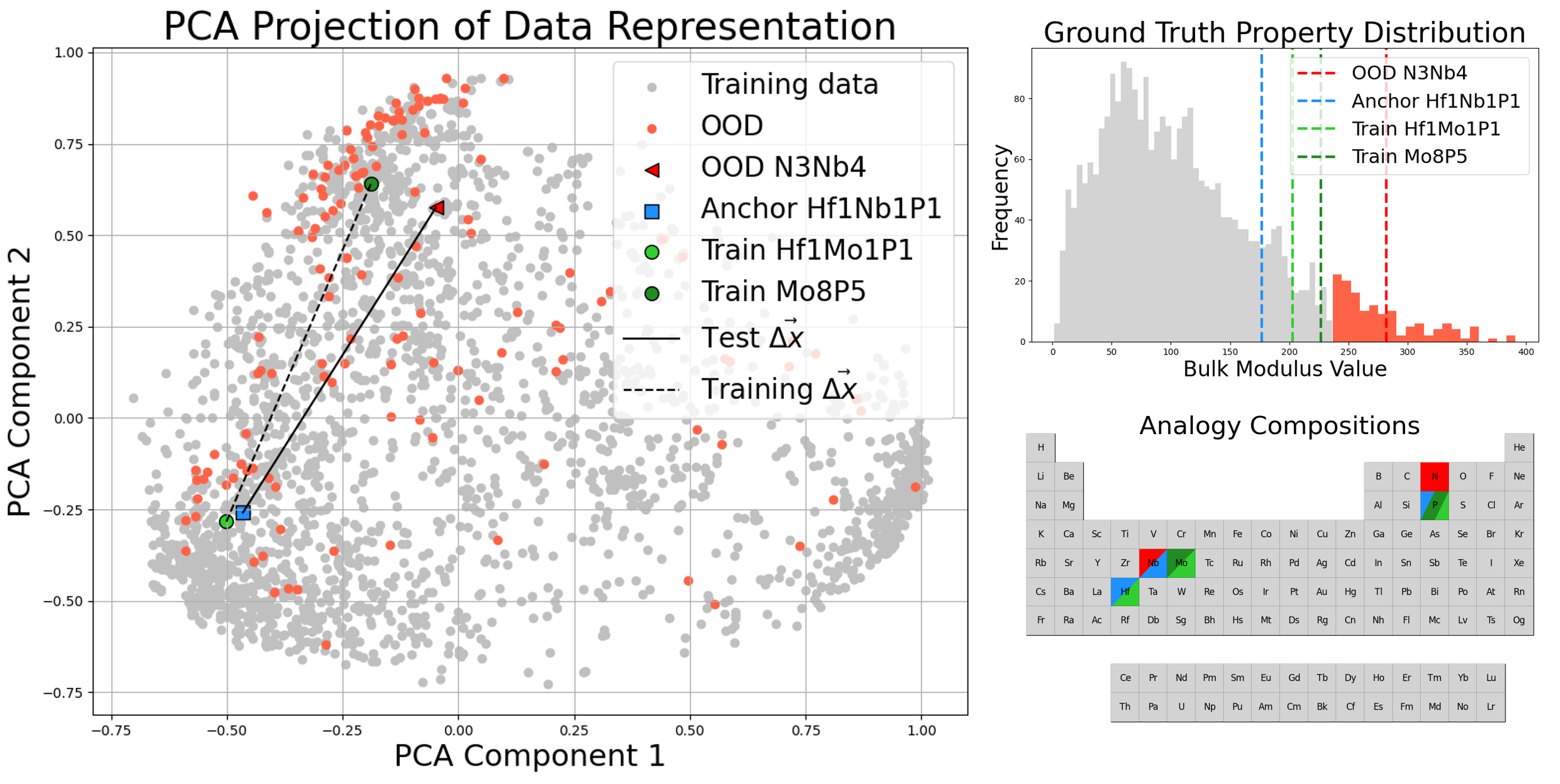}
    \caption{\textbf{Visualizing Analogies in Bilinear Transduction}. AFLOW bulk modulus \ood predictions are based on in-distribution anchors, that paired with \ood points, form analogies to training pairs.
    (a) PCA plot of all samples in the dataset. The difference between the \ood point and its anchor is similar to the difference between the training point and anchor. 
    (b) Ground truth shear modulus training (gray) and test (red) distributions and \ood, anchor, and analogous training pair values.
    (c) test \textcolor{Cerulean}{anchor} and training \textcolor{LimeGreen}{anchor} differ by one neighboring d-block element (\ce{Mo}-\ce{Nb}). \textcolor{red}{\ood} and training \textcolor{ForestGreen}{target} differ by the same elements, and in a group 15 element (\ce{N}-\ce{P}).
    }
    \label{fig:appendix_analogies2}
\end{figure*}

\paragraph{Solids}
\label{subsubssec:appendix_analogies_solids}
Figure~\ref{fig:appendix_analogies1} describes shear modulus inference of \ood \ce{NOs} via anchor \ce{NIr}, analogous to training anchor \ce{CaP_{2}Rh_{2}} and training target \ce{CaP_{2}Ru_{2}}. In this case, the training anchor and target (\ce{CaP_{2}Rh_{2}} and \ce{CaP_{2}Ru_{2}}) and test anchor and target (\ce{NIr} and \ce{NOs}) differ by one d-block element: \ce{Rh (Z=45)} to \ce{Ru (Z=44)} and \ce{Ir (Z=77)} to \ce{Os (Z=76)}. Figure~\ref{fig:appendix_analogies2} describes bulk modulus inference of \ood \ce{N_{3}Nb_{4}} via anchor \ce{HfNbP}, analogous to training anchor \ce{HfMoP} and training target \ce{Mo_{8}P_{5}}. 
Additional examples include shear modulus prediction for \ood \ce{NbSiIr} via anchor \ce{NbSiPt} analogous to training anchor \ce{GeNbIr} and target \ce{GeNbPt}, and \ood \ce{ReSiNb} via anchor \ce{OsSiZr} analogous to training anchor \ce{GeIrNb} and target \ce{GeIrPt}.

\begin{figure*}[htbp]
    \centering
    \includegraphics[width=0.99\linewidth]{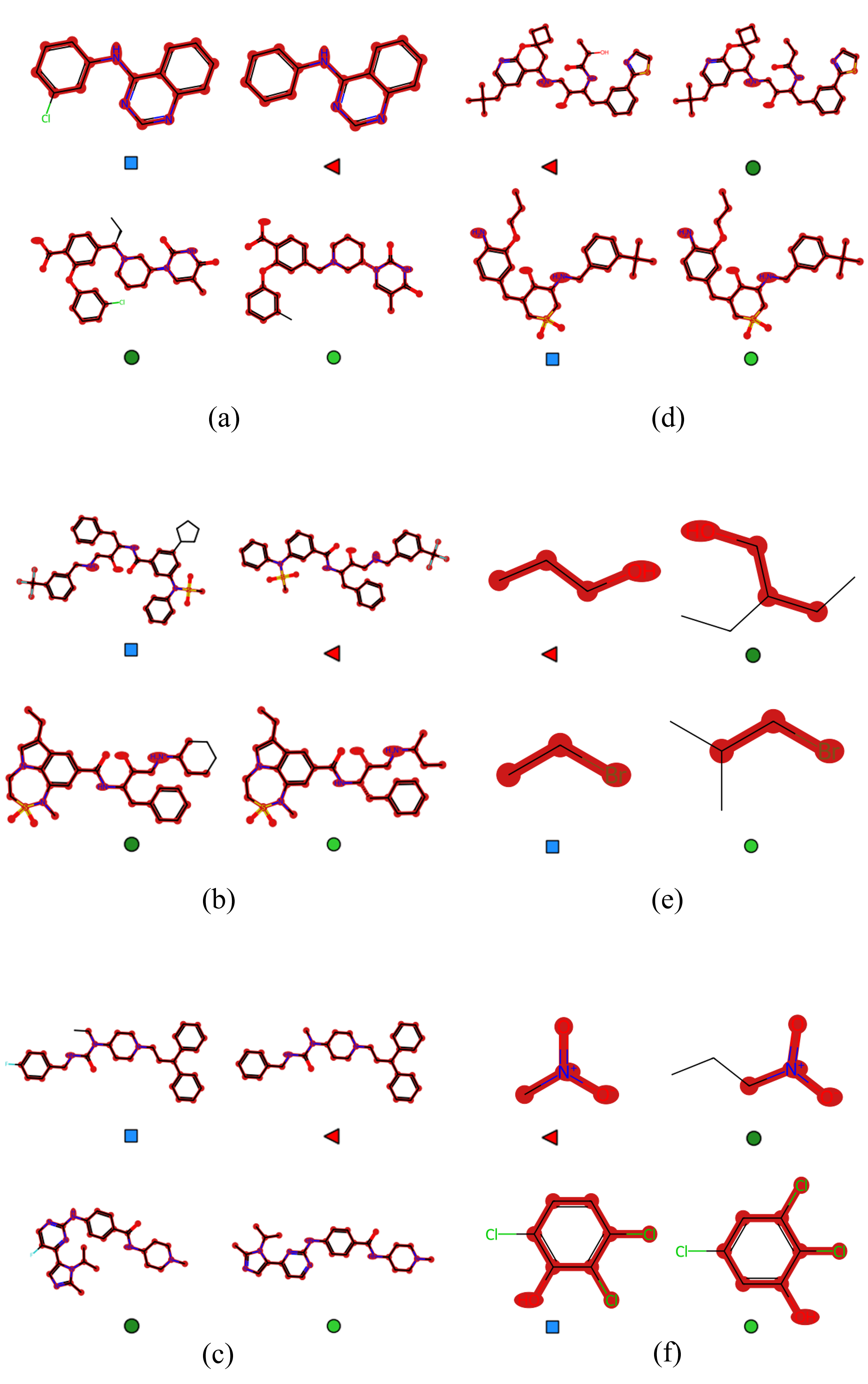}
    \caption{\textbf{Analogical Molecules pairs.} 
    Molecules are paired according to maximum similarity. (a-c) \textbf{top} pair: \textcolor{red}{\ood} are similar to their \textcolor{Cerulean}{anchors}, \textbf{bottom} pair: training \textcolor{ForestGreen}{targets} are similar to their \textcolor{LimeGreen}{anchors}.
    (d-f) \textbf{top} pair: \textcolor{red}{\ood} are similar to training \textcolor{ForestGreen}{targets}, \textbf{bottom} pair: \textcolor{Cerulean}{ anchors} are similar to training \textcolor{LimeGreen}{anchors}. For each pair, we denote the relevant benchmark and chemical operator differentiating samples within each pair. (a) lipophilicity, \ce{Cl} addition. (b) BACE, addition or completion of a ring. (c) lipophilicity, \ce{F} addition. (d) BACE, targets differ in \ce{OH} functional group and anchors are the same. (e) ESOL, additional \ce{C}. (f) ESOL, targets differ in additional \ce{C} and anchors differ in functional group position in the ring.
    }
    \label{fig:appendix_analogies_only_mols}
\end{figure*}

\paragraph{Molecules}
\label{subsubssec:appendix_analogies_mol}
Figures~\ref{fig:analogies_mol} and~\ref{fig:appendix_analogies_only_mols} display additional analogical molecule pairs. Two distinct modes of similarity can be identified: one is between the training anchor and target, and between the test anchor and target Figures~\ref{fig:appendix_analogies_only_mols}\textcolor{blue}{(a,b,c)}, and the other is between the anchors, and between the targets Figure~\ref{fig:appendix_analogies_only_mols}\textcolor{blue}{(d,e,f)}. In the anchor selection process, the model can converge to an anchor that is either very similar to the \ood target, in a way that two training samples are similar (in $\mathcal{X}$ space), or that is different from the \ood target and similar to the training anchor and the training target will be similar to the \ood. For both modes, we can spot analogous differences between the molecules.

\begin{figure*}[htbp]
    \centering
    \includegraphics[width=0.99\linewidth]{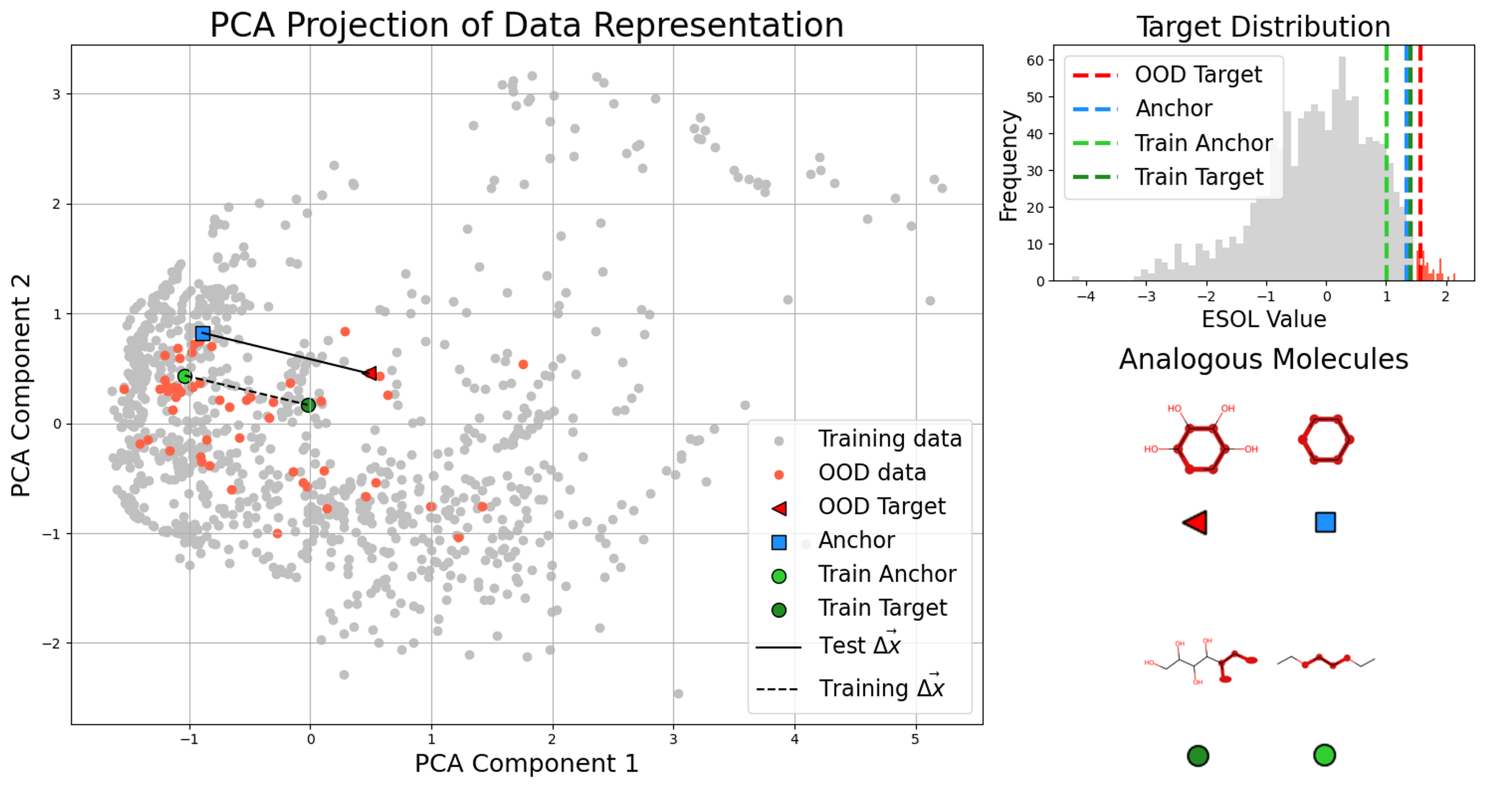}
    \caption{\textbf{Visualizing Analogies in Bilinear Transduction}. MoleculeNet ESOL \ood predictions are based on in-distribution anchors, that paired with \ood points, form analogies to training pairs.
    (a) PCA plot of all samples in the dataset. The difference between the \ood point and its anchor is similar to the difference between the training point and anchor.
    (b) Ground truth ESOL training (gray) and test (red) distributions and \ood, anchor, and analogous training pair values.
    (c) Analogical molecules pairs. The similarity between \textcolor{red}{\ood}-\textcolor{Cerulean}{anchor} and training \textcolor{LimeGreen}{anchor} and \textcolor{ForestGreen}{target} is highlighted in red, using the MCS metric.
    }
    \label{fig:appendix_analogies_mol}
\end{figure*}

\clearpage
\subsection{Implementation Details}\label{appendix-details}

\subsubsection{Data Representation}
\label{appendix:rep}

\textbf{Chemprop.} Chemprop derives a molecular graph representation based on SMILES, featurizes the atom nodes and bond edges using RDKit, then preforms message-passing in order to learn an intermediate representation for property prediction via a feed-forward network \cite{weininger1988smiles, rdkit, heid2023chemprop}.

\textbf{RF} and \textbf{MLP.} We derive RDKit descriptors from the SMILES representation using the deepchem library \cite{rdkit, weininger1988smiles, ramsundar2018molecular}.

\textbf{Ridge Regression.} We adhere to the data preprocessing scheme outlined in \citet{kauwe2020can} on AFLOW and featurize data using element-based Oliynyk descriptors \cite{oliynyk2016high}.  
The data is scaled using the StandardScaler and normalized using the Normalizer from the sklearn library based on the training data statistics.

\textbf{MODNet.} Following \citet{de2021materials} on Matbench, we create element-based feature vectors using Matminer \cite{ward2018matminer}, and then select features based on the normalized mutual information \cite{kraskov2004estimating}. The data is scaled using the MinMaxScaler based on the training data statistics and NaN values are replaced with a constant, -1.

\textbf{CrabNet.} 
Following \citet{wang2021compositionally} on MP, we leverage mat2vec, learned via self-supervised natural language processing techniques, trained on a large corpus of scientific literature.

\textbf{Bilinear Transduction.}
For MoleculeNet, we derive RDKit descriptors from the SMILES representation using the deepchem library \cite{rdkit, weininger1988smiles, ramsundar2018molecular}. 
We use the Ridge Regression representation and processing for AFLOW and MP, and the MODNet representation and processing for Matbench.

\subsubsection{Bilinear Transduction Hyperparameter Search}

In Table~\ref{tab:analysis}, we report the best \ood MAE score for Bilinear Transduction on AFLOW over a hyperparameter search on the number of predictor network layers ($3,4$), layer size ($256,512,1024$) and embedding size ($32,42,48,64$).
The hyperparameter search revealed little sensitivity to changes in hyperparameter values, indicating the robustness of our evaluation.

\clearpage
\subsection{Methods}\label{appendix-blt}
\paragraph{Bilinear Transduction}
Bilinear Transduction \cite{netanyahu2023transduction} is a regression method for out-of-support (\oos) generalization, a type of distribution shift where data becomes \oos at test time.
By reparameterizing the problem, \oos generalization is possible via a bilinear predictor.
Let $\mathcal{X}\subseteq\mathbb{R}^n$ denote a representation space of data points and let $\mathcal{Y}\subseteq\mathbb{R}$ denote the space of labels. Given a training set $\mathcal{D}^{\mathrm{tr}}=\{(x,y)\}_n\subseteq\mathcal{X}\times\mathcal{Y}$, Bilinear Transduction learns a transductive predictor that extrapolates in a zero-shot manner to \oos test points. 
Naively, one can learn predictor $h_{\theta}:\mathcal{X} \to \mathcal{Y}$. However, ML methods often fail under covariate shift \cite{shimodaira2000improving,DatasetShiftBook}.
Instead, Bilinear Transduction offers the following formulation.

Let $\Delta \mathcal{X}=\{x_i-x_j|x_i,x_j\in\mathcal{X}\}$ denote the differences distribution. During training, predictor $h_{\theta}:\Delta \mathcal{X}\times\mathcal{X}\to\mathcal{Y}$ predicts value $y_2\in\mathcal{D}^{\mathrm{tr}}_\mathcal{Y}$ for data point $x_2\in\mathcal{D}^{\mathrm{tr}}_\mathcal{X}$ given anchor point $x_1\in\mathcal{D}^{\mathrm{tr}}_\mathcal{X}$ and the difference between them $\Delta x_{21}=x_2-x_1\in\mathcal{D}^{\mathrm{tr}}_{\Delta \mathcal{X}}$. 
The predictor $h_{\theta}(\Delta x,x)=f_{\theta}(\Delta x)g_{\theta}(x)$ is implemented as a bilinear function in non-linear embeddings of $\Delta x$ and $x$.

The test set $\mathcal{D}^{\mathrm{te}}\subseteq\mathcal{X}$ includes \oos data points. 
Given test point $x_{\mathrm{te}}\in\mathcal{D}^{\mathrm{te}}$, its predicted value is $h_{\theta}(\Delta x_{\mathrm{te,an}},x_{\mathrm{an}})$. Anchor $x_{\mathrm{an}}\in\mathcal{D}^{\mathrm{tr}}_\mathcal{X}$ is the training point that minimizes the distance between its difference with the test point $\Delta x_{\mathrm{te,an}}=x_{\mathrm{te}}-x_{\mathrm{an}}$, and differences within the training distribution $\mathcal{D}^{\mathrm{tr}}_{\Delta X}$. Formally, $x_{\mathrm{an}}=\mathrm{argmin}_{x_i\in\mathcal{D}^{\mathrm{tr}}}\{||\Delta x_{\mathrm{te},i}-\Delta x||_2\mid \Delta x\in\mathcal{D}^{\mathrm{tr}}_{\Delta X}\}$.
This reparameterization converts the problem to within support, as $\Delta x_{\mathrm{te,an}}$ and $x_{\mathrm{an}}$ are within the training distribution.
Under some coverage and rank assumptions Bilinear Transduction has theoretical convergence guarantees \cite{netanyahu2023transduction}.

We adopt Bilinear Transduction for out setting where the output $\mathcal{Y}$ is \ood rather than the input being \oos (described in Section~\ref{methods}). 
Under our current formulation where $\mathcal{Y}$ is \ood, the theoretical guarantees for Bilinear Transduction may not fully apply. Future work will focus on learning data representations that align with the assumptions of Bilinear Transduction for \ood $\mathcal{X}$.

\clearpage
\bibliography{sn-bibliography}


\begin{thebibliography}{41}
\ifx \bisbn   \undefined \def \bisbn  #1{ISBN #1}\fi
\ifx \binits  \undefined \def \binits#1{#1}\fi
\ifx \bauthor  \undefined \def \bauthor#1{#1}\fi
\ifx \batitle  \undefined \def \batitle#1{#1}\fi
\ifx \bjtitle  \undefined \def \bjtitle#1{#1}\fi
\ifx \bvolume  \undefined \def \bvolume#1{\textbf{#1}}\fi
\ifx \byear  \undefined \def \byear#1{#1}\fi
\ifx \bissue  \undefined \def \bissue#1{#1}\fi
\ifx \bfpage  \undefined \def \bfpage#1{#1}\fi
\ifx \blpage  \undefined \def \blpage #1{#1}\fi
\ifx \burl  \undefined \def \burl#1{\textsf{#1}}\fi
\ifx \doiurl  \undefined \def \doiurl#1{\url{https://doi.org/#1}}\fi
\ifx \betal  \undefined \def \betal{\textit{et al.}}\fi
\ifx \binstitute  \undefined \def \binstitute#1{#1}\fi
\ifx \binstitutionaled  \undefined \def \binstitutionaled#1{#1}\fi
\ifx \bctitle  \undefined \def \bctitle#1{#1}\fi
\ifx \beditor  \undefined \def \beditor#1{#1}\fi
\ifx \bpublisher  \undefined \def \bpublisher#1{#1}\fi
\ifx \bbtitle  \undefined \def \bbtitle#1{#1}\fi
\ifx \bedition  \undefined \def \bedition#1{#1}\fi
\ifx \bseriesno  \undefined \def \bseriesno#1{#1}\fi
\ifx \blocation  \undefined \def \blocation#1{#1}\fi
\ifx \bsertitle  \undefined \def \bsertitle#1{#1}\fi
\ifx \bsnm \undefined \def \bsnm#1{#1}\fi
\ifx \bsuffix \undefined \def \bsuffix#1{#1}\fi
\ifx \bparticle \undefined \def \bparticle#1{#1}\fi
\ifx \barticle \undefined \def \barticle#1{#1}\fi
\bibcommenthead
\ifx \bconfdate \undefined \def \bconfdate #1{#1}\fi
\ifx \botherref \undefined \def \botherref #1{#1}\fi
\ifx \url \undefined \def \url#1{\textsf{#1}}\fi
\ifx \bchapter \undefined \def \bchapter#1{#1}\fi
\ifx \bbook \undefined \def \bbook#1{#1}\fi
\ifx \bcomment \undefined \def \bcomment#1{#1}\fi
\ifx \oauthor \undefined \def \oauthor#1{#1}\fi
\ifx \citeauthoryear \undefined \def \citeauthoryear#1{#1}\fi
\ifx \endbibitem  \undefined \def \endbibitem {}\fi
\ifx \bconflocation  \undefined \def \bconflocation#1{#1}\fi
\ifx \arxivurl  \undefined \def \arxivurl#1{\textsf{#1}}\fi
\csname PreBibitemsHook\endcsname

%%% 1
\bibitem[\protect\citeauthoryear{Axelrod et~al.}{2022}]{axelrod2022learning}
\begin{barticle}
\bauthor{\bsnm{Axelrod}, \binits{S.}},
\bauthor{\bsnm{Schwalbe-Koda}, \binits{D.}},
\bauthor{\bsnm{Mohapatra}, \binits{S.}},
\bauthor{\bsnm{Damewood}, \binits{J.}},
\bauthor{\bsnm{Greenman}, \binits{K.P.}},
\bauthor{\bsnm{G{\'o}mez-Bombarelli}, \binits{R.}}:
\batitle{Learning matter: Materials design with machine learning and atomistic simulations}.
\bjtitle{Accounts of Materials Research}
\bvolume{3}(\bissue{3}),
\bfpage{343}--\blpage{357}
(\byear{2022})
\end{barticle}
\endbibitem

%%% 2
\bibitem[\protect\citeauthoryear{Sanchez-Lengeling and Aspuru-Guzik}{2018}]{sanchez2018inverse}
\begin{barticle}
\bauthor{\bsnm{Sanchez-Lengeling}, \binits{B.}},
\bauthor{\bsnm{Aspuru-Guzik}, \binits{A.}}:
\batitle{Inverse molecular design using machine learning: Generative models for matter engineering}.
\bjtitle{Science}
\bvolume{361}(\bissue{6400}),
\bfpage{360}--\blpage{365}
(\byear{2018})
\end{barticle}
\endbibitem

%%% 3
\bibitem[\protect\citeauthoryear{Bilodeau et~al.}{2022}]{bilodeau2022generative}
\begin{barticle}
\bauthor{\bsnm{Bilodeau}, \binits{C.}},
\bauthor{\bsnm{Jin}, \binits{W.}},
\bauthor{\bsnm{Jaakkola}, \binits{T.}},
\bauthor{\bsnm{Barzilay}, \binits{R.}},
\bauthor{\bsnm{Jensen}, \binits{K.F.}}:
\batitle{Generative models for molecular discovery: Recent advances and challenges}.
\bjtitle{Wiley Interdisciplinary Reviews: Computational Molecular Science}
\bvolume{12}(\bissue{5}),
\bfpage{1608}
(\byear{2022})
\end{barticle}
\endbibitem

%%% 4
\bibitem[\protect\citeauthoryear{Noh et~al.}{2020}]{noh2020machine}
\begin{barticle}
\bauthor{\bsnm{Noh}, \binits{J.}},
\bauthor{\bsnm{Gu}, \binits{G.H.}},
\bauthor{\bsnm{Kim}, \binits{S.}},
\bauthor{\bsnm{Jung}, \binits{Y.}}:
\batitle{Machine-enabled inverse design of inorganic solid materials: promises and challenges}.
\bjtitle{Chemical Science}
\bvolume{11}(\bissue{19}),
\bfpage{4871}--\blpage{4881}
(\byear{2020})
\end{barticle}
\endbibitem

%%% 5
\bibitem[\protect\citeauthoryear{Kim et~al.}{2021}]{kim2021deep}
\begin{barticle}
\bauthor{\bsnm{Kim}, \binits{Y.}},
\bauthor{\bsnm{Kim}, \binits{Y.}},
\bauthor{\bsnm{Yang}, \binits{C.}},
\bauthor{\bsnm{Park}, \binits{K.}},
\bauthor{\bsnm{Gu}, \binits{G.X.}},
\bauthor{\bsnm{Ryu}, \binits{S.}}:
\batitle{Deep learning framework for material design space exploration using active transfer learning and data augmentation}.
\bjtitle{npj Computational Materials}
\bvolume{7}(\bissue{1}),
\bfpage{140}
(\byear{2021})
\end{barticle}
\endbibitem

%%% 6
\bibitem[\protect\citeauthoryear{Zeni et~al.}{2023}]{zeni2023mattergen}
\begin{botherref}
\oauthor{\bsnm{Zeni}, \binits{C.}},
\oauthor{\bsnm{Pinsler}, \binits{R.}},
\oauthor{\bsnm{Z{\"u}gner}, \binits{D.}},
\oauthor{\bsnm{Fowler}, \binits{A.}},
\oauthor{\bsnm{Horton}, \binits{M.}},
\oauthor{\bsnm{Fu}, \binits{X.}},
\oauthor{\bsnm{Shysheya}, \binits{S.}},
\oauthor{\bsnm{Crabb{\'e}}, \binits{J.}},
\oauthor{\bsnm{Sun}, \binits{L.}},
\oauthor{\bsnm{Smith}, \binits{J.}}, et al.:
Mattergen: a generative model for inorganic materials design.
arXiv preprint arXiv:2312.03687
(2023)
\end{botherref}
\endbibitem

%%% 7
\bibitem[\protect\citeauthoryear{Yang et~al.}{2023}]{yang2023scalable}
\begin{botherref}
\oauthor{\bsnm{Yang}, \binits{M.}},
\oauthor{\bsnm{Cho}, \binits{K.}},
\oauthor{\bsnm{Merchant}, \binits{A.}},
\oauthor{\bsnm{Abbeel}, \binits{P.}},
\oauthor{\bsnm{Schuurmans}, \binits{D.}},
\oauthor{\bsnm{Mordatch}, \binits{I.}},
\oauthor{\bsnm{Cubuk}, \binits{E.D.}}:
Scalable diffusion for materials generation.
arXiv preprint arXiv:2311.09235
(2023)
\end{botherref}
\endbibitem

%%% 8
\bibitem[\protect\citeauthoryear{Xie et~al.}{2021}]{xie2021crystal}
\begin{botherref}
\oauthor{\bsnm{Xie}, \binits{T.}},
\oauthor{\bsnm{Fu}, \binits{X.}},
\oauthor{\bsnm{Ganea}, \binits{O.-E.}},
\oauthor{\bsnm{Barzilay}, \binits{R.}},
\oauthor{\bsnm{Jaakkola}, \binits{T.}}:
Crystal diffusion variational autoencoder for periodic material generation.
arXiv preprint arXiv:2110.06197
(2021)
\end{botherref}
\endbibitem

%%% 9
\bibitem[\protect\citeauthoryear{Walters and Barzilay}{2020}]{walters2020applications}
\begin{barticle}
\bauthor{\bsnm{Walters}, \binits{W.P.}},
\bauthor{\bsnm{Barzilay}, \binits{R.}}:
\batitle{Applications of deep learning in molecule generation and molecular property prediction}.
\bjtitle{Accounts of chemical research}
\bvolume{54}(\bissue{2}),
\bfpage{263}--\blpage{270}
(\byear{2020})
\end{barticle}
\endbibitem

%%% 10
\bibitem[\protect\citeauthoryear{Dunn et~al.}{2020}]{dunn2020benchmarking}
\begin{barticle}
\bauthor{\bsnm{Dunn}, \binits{A.}},
\bauthor{\bsnm{Wang}, \binits{Q.}},
\bauthor{\bsnm{Ganose}, \binits{A.}},
\bauthor{\bsnm{Dopp}, \binits{D.}},
\bauthor{\bsnm{Jain}, \binits{A.}}:
\batitle{Benchmarking materials property prediction methods: the matbench test set and automatminer reference algorithm}.
\bjtitle{npj Computational Materials}
\bvolume{6}(\bissue{1}),
\bfpage{138}
(\byear{2020})
\end{barticle}
\endbibitem

%%% 11
\bibitem[\protect\citeauthoryear{Wang et~al.}{2021}]{wang2021compositionally}
\begin{barticle}
\bauthor{\bsnm{Wang}, \binits{A.Y.-T.}},
\bauthor{\bsnm{Kauwe}, \binits{S.K.}},
\bauthor{\bsnm{Murdock}, \binits{R.J.}},
\bauthor{\bsnm{Sparks}, \binits{T.D.}}:
\batitle{Compositionally restricted attention-based network for materials property predictions}.
\bjtitle{Npj Computational Materials}
\bvolume{7}(\bissue{1}),
\bfpage{77}
(\byear{2021})
\end{barticle}
\endbibitem

%%% 12
\bibitem[\protect\citeauthoryear{Zhuo et~al.}{2018}]{zhuo2018predicting}
\begin{barticle}
\bauthor{\bsnm{Zhuo}, \binits{Y.}},
\bauthor{\bsnm{Mansouri~Tehrani}, \binits{A.}},
\bauthor{\bsnm{Brgoch}, \binits{J.}}:
\batitle{Predicting the band gaps of inorganic solids by machine learning}.
\bjtitle{The journal of physical chemistry letters}
\bvolume{9}(\bissue{7}),
\bfpage{1668}--\blpage{1673}
(\byear{2018})
\end{barticle}
\endbibitem

%%% 13
\bibitem[\protect\citeauthoryear{De~Breuck et~al.}{2021}]{de2021materials}
\begin{barticle}
\bauthor{\bsnm{De~Breuck}, \binits{P.-P.}},
\bauthor{\bsnm{Hautier}, \binits{G.}},
\bauthor{\bsnm{Rignanese}, \binits{G.-M.}}:
\batitle{Materials property prediction for limited datasets enabled by feature selection and joint learning with modnet}.
\bjtitle{npj computational materials}
\bvolume{7}(\bissue{1}),
\bfpage{83}
(\byear{2021})
\end{barticle}
\endbibitem

%%% 14
\bibitem[\protect\citeauthoryear{Ward et~al.}{2016}]{ward2016general}
\begin{barticle}
\bauthor{\bsnm{Ward}, \binits{L.}},
\bauthor{\bsnm{Agrawal}, \binits{A.}},
\bauthor{\bsnm{Choudhary}, \binits{A.}},
\bauthor{\bsnm{Wolverton}, \binits{C.}}:
\batitle{A general-purpose machine learning framework for predicting properties of inorganic materials}.
\bjtitle{npj Computational Materials}
\bvolume{2}(\bissue{1}),
\bfpage{1}--\blpage{7}
(\byear{2016})
\end{barticle}
\endbibitem

%%% 15
\bibitem[\protect\citeauthoryear{Kauwe et~al.}{2020}]{kauwe2020can}
\begin{barticle}
\bauthor{\bsnm{Kauwe}, \binits{S.K.}},
\bauthor{\bsnm{Graser}, \binits{J.}},
\bauthor{\bsnm{Murdock}, \binits{R.}},
\bauthor{\bsnm{Sparks}, \binits{T.D.}}:
\batitle{Can machine learning find extraordinary materials?}
\bjtitle{Computational Materials Science}
\bvolume{174},
\bfpage{109498}
(\byear{2020})
\end{barticle}
\endbibitem

%%% 16
\bibitem[\protect\citeauthoryear{Zhao et~al.}{2022}]{zhao2022limitations}
\begin{barticle}
\bauthor{\bsnm{Zhao}, \binits{Z.-W.}},
\bauthor{\bsnm{Del~Cueto}, \binits{M.}},
\bauthor{\bsnm{Troisi}, \binits{A.}}:
\batitle{Limitations of machine learning models when predicting compounds with completely new chemistries: possible improvements applied to the discovery of new non-fullerene acceptors}.
\bjtitle{Digital Discovery}
\bvolume{1}(\bissue{3}),
\bfpage{266}--\blpage{276}
(\byear{2022})
\end{barticle}
\endbibitem

%%% 17
\bibitem[\protect\citeauthoryear{Omee et~al.}{2024}]{omee2024structure}
\begin{barticle}
\bauthor{\bsnm{Omee}, \binits{S.S.}},
\bauthor{\bsnm{Fu}, \binits{N.}},
\bauthor{\bsnm{Dong}, \binits{R.}},
\bauthor{\bsnm{Hu}, \binits{M.}},
\bauthor{\bsnm{Hu}, \binits{J.}}:
\batitle{Structure-based out-of-distribution (ood) materials property prediction: a benchmark study}.
\bjtitle{npj Computational Materials}
\bvolume{10}(\bissue{1}),
\bfpage{144}
(\byear{2024})
\end{barticle}
\endbibitem

%%% 18
\bibitem[\protect\citeauthoryear{Li et~al.}{2024}]{li2024probing}
\begin{botherref}
\oauthor{\bsnm{Li}, \binits{K.}},
\oauthor{\bsnm{Rubungo}, \binits{A.N.}},
\oauthor{\bsnm{Lei}, \binits{X.}},
\oauthor{\bsnm{Persaud}, \binits{D.}},
\oauthor{\bsnm{Choudhary}, \binits{K.}},
\oauthor{\bsnm{DeCost}, \binits{B.}},
\oauthor{\bsnm{Dieng}, \binits{A.B.}},
\oauthor{\bsnm{Hattrick-Simpers}, \binits{J.}}:
Probing out-of-distribution generalization in machine learning for materials.
arXiv preprint arXiv:2406.06489
(2024)
\end{botherref}
\endbibitem

%%% 19
\bibitem[\protect\citeauthoryear{Meredig et~al.}{2018}]{meredig2018can}
\begin{barticle}
\bauthor{\bsnm{Meredig}, \binits{B.}},
\bauthor{\bsnm{Antono}, \binits{E.}},
\bauthor{\bsnm{Church}, \binits{C.}},
\bauthor{\bsnm{Hutchinson}, \binits{M.}},
\bauthor{\bsnm{Ling}, \binits{J.}},
\bauthor{\bsnm{Paradiso}, \binits{S.}},
\bauthor{\bsnm{Blaiszik}, \binits{B.}},
\bauthor{\bsnm{Foster}, \binits{I.}},
\bauthor{\bsnm{Gibbons}, \binits{B.}},
\bauthor{\bsnm{Hattrick-Simpers}, \binits{J.}}, \betal:
\batitle{Can machine learning identify the next high-temperature superconductor? examining extrapolation performance for materials discovery}.
\bjtitle{Molecular Systems Design \& Engineering}
\bvolume{3}(\bissue{5}),
\bfpage{819}--\blpage{825}
(\byear{2018})
\end{barticle}
\endbibitem

%%% 20
\bibitem[\protect\citeauthoryear{Muckley et~al.}{2023}]{muckley2023interpretable}
\begin{barticle}
\bauthor{\bsnm{Muckley}, \binits{E.S.}},
\bauthor{\bsnm{Saal}, \binits{J.E.}},
\bauthor{\bsnm{Meredig}, \binits{B.}},
\bauthor{\bsnm{Roper}, \binits{C.S.}},
\bauthor{\bsnm{Martin}, \binits{J.H.}}:
\batitle{Interpretable models for extrapolation in scientific machine learning}.
\bjtitle{Digital Discovery}
\bvolume{2}(\bissue{5}),
\bfpage{1425}--\blpage{1435}
(\byear{2023})
\end{barticle}
\endbibitem

%%% 21
\bibitem[\protect\citeauthoryear{Yang et~al.}{2024}]{yang2024mattersim}
\begin{botherref}
\oauthor{\bsnm{Yang}, \binits{H.}},
\oauthor{\bsnm{Hu}, \binits{C.}},
\oauthor{\bsnm{Zhou}, \binits{Y.}},
\oauthor{\bsnm{Liu}, \binits{X.}},
\oauthor{\bsnm{Shi}, \binits{Y.}},
\oauthor{\bsnm{Li}, \binits{J.}},
\oauthor{\bsnm{Li}, \binits{G.}},
\oauthor{\bsnm{Chen}, \binits{Z.}},
\oauthor{\bsnm{Chen}, \binits{S.}},
\oauthor{\bsnm{Zeni}, \binits{C.}}, et al.:
Mattersim: A deep learning atomistic model across elements, temperatures and pressures.
arXiv preprint arXiv:2405.04967
(2024)
\end{botherref}
\endbibitem

%%% 22
\bibitem[\protect\citeauthoryear{Merchant et~al.}{2023}]{merchant2023scaling}
\begin{barticle}
\bauthor{\bsnm{Merchant}, \binits{A.}},
\bauthor{\bsnm{Batzner}, \binits{S.}},
\bauthor{\bsnm{Schoenholz}, \binits{S.S.}},
\bauthor{\bsnm{Aykol}, \binits{M.}},
\bauthor{\bsnm{Cheon}, \binits{G.}},
\bauthor{\bsnm{Cubuk}, \binits{E.D.}}:
\batitle{Scaling deep learning for materials discovery}.
\bjtitle{Nature}
\bvolume{624}(\bissue{7990}),
\bfpage{80}--\blpage{85}
(\byear{2023})
\end{barticle}
\endbibitem

%%% 23
\bibitem[\protect\citeauthoryear{Batatia et~al.}{2023}]{batatia2023foundation}
\begin{botherref}
\oauthor{\bsnm{Batatia}, \binits{I.}},
\oauthor{\bsnm{Benner}, \binits{P.}},
\oauthor{\bsnm{Chiang}, \binits{Y.}},
\oauthor{\bsnm{Elena}, \binits{A.M.}},
\oauthor{\bsnm{Kov{\'a}cs}, \binits{D.P.}},
\oauthor{\bsnm{Riebesell}, \binits{J.}},
\oauthor{\bsnm{Advincula}, \binits{X.R.}},
\oauthor{\bsnm{Asta}, \binits{M.}},
\oauthor{\bsnm{Baldwin}, \binits{W.J.}},
\oauthor{\bsnm{Bernstein}, \binits{N.}}, et al.:
A foundation model for atomistic materials chemistry.
arXiv preprint arXiv:2401.00096
(2023)
\end{botherref}
\endbibitem

%%% 24
\bibitem[\protect\citeauthoryear{Netanyahu et~al.}{2023}]{netanyahu2023transduction}
\begin{bchapter}
\bauthor{\bsnm{Netanyahu}, \binits{A.}},
\bauthor{\bsnm{Gupta}, \binits{A.}},
\bauthor{\bsnm{Simchowitz}, \binits{M.}},
\bauthor{\bsnm{Zhang}, \binits{K.}},
\bauthor{\bsnm{Agrawal}, \binits{P.}}:
\bctitle{Learning to extrapolate: A transductive approach}.
In: \bbtitle{International Conference on Learning Representations}
(\byear{2023})
\end{bchapter}
\endbibitem

%%% 25
\bibitem[\protect\citeauthoryear{Curtarolo et~al.}{2012}]{curtarolo2012aflow}
\begin{barticle}
\bauthor{\bsnm{Curtarolo}, \binits{S.}},
\bauthor{\bsnm{Setyawan}, \binits{W.}},
\bauthor{\bsnm{Hart}, \binits{G.L.}},
\bauthor{\bsnm{Jahnatek}, \binits{M.}},
\bauthor{\bsnm{Chepulskii}, \binits{R.V.}},
\bauthor{\bsnm{Taylor}, \binits{R.H.}},
\bauthor{\bsnm{Wang}, \binits{S.}},
\bauthor{\bsnm{Xue}, \binits{J.}},
\bauthor{\bsnm{Yang}, \binits{K.}},
\bauthor{\bsnm{Levy}, \binits{O.}}, \betal:
\batitle{Aflow: An automatic framework for high-throughput materials discovery}.
\bjtitle{Computational Materials Science}
\bvolume{58},
\bfpage{218}--\blpage{226}
(\byear{2012})
\end{barticle}
\endbibitem

%%% 26
\bibitem[\protect\citeauthoryear{Jain et~al.}{2013}]{jain2013commentary}
\begin{botherref}
\oauthor{\bsnm{Jain}, \binits{A.}},
\oauthor{\bsnm{Ong}, \binits{S.P.}},
\oauthor{\bsnm{Hautier}, \binits{G.}},
\oauthor{\bsnm{Chen}, \binits{W.}},
\oauthor{\bsnm{Richards}, \binits{W.D.}},
\oauthor{\bsnm{Dacek}, \binits{S.}},
\oauthor{\bsnm{Cholia}, \binits{S.}},
\oauthor{\bsnm{Gunter}, \binits{D.}},
\oauthor{\bsnm{Skinner}, \binits{D.}},
\oauthor{\bsnm{Ceder}, \binits{G.}}, et al.:
Commentary: The materials project: A materials genome approach to accelerating materials innovation.
APL materials
\textbf{1}(1)
(2013)
\end{botherref}
\endbibitem

%%% 27
\bibitem[\protect\citeauthoryear{G. and S.}{2017}]{CiteDrive2022}
\begin{botherref}
\oauthor{\bsnm{G.}, \binits{C.}},
\oauthor{\bsnm{S.}, \binits{B.}}:
Mechanical properties of some steels.
\url{https://citrination.com/datasets/153092/show_files/}
(2017)
\end{botherref}
\endbibitem

%%% 28
\bibitem[\protect\citeauthoryear{Petousis et~al.}{2017}]{petousis2017high}
\begin{barticle}
\bauthor{\bsnm{Petousis}, \binits{I.}},
\bauthor{\bsnm{Mrdjenovich}, \binits{D.}},
\bauthor{\bsnm{Ballouz}, \binits{E.}},
\bauthor{\bsnm{Liu}, \binits{M.}},
\bauthor{\bsnm{Winston}, \binits{D.}},
\bauthor{\bsnm{Chen}, \binits{W.}},
\bauthor{\bsnm{Graf}, \binits{T.}},
\bauthor{\bsnm{Schladt}, \binits{T.D.}},
\bauthor{\bsnm{Persson}, \binits{K.A.}},
\bauthor{\bsnm{Prinz}, \binits{F.B.}}:
\batitle{High-throughput screening of inorganic compounds for the discovery of novel dielectric and optical materials}.
\bjtitle{Scientific data}
\bvolume{4}(\bissue{1}),
\bfpage{1}--\blpage{12}
(\byear{2017})
\end{barticle}
\endbibitem

%%% 29
\bibitem[\protect\citeauthoryear{Heid et~al.}{2023}]{heid2023chemprop}
\begin{barticle}
\bauthor{\bsnm{Heid}, \binits{E.}},
\bauthor{\bsnm{Greenman}, \binits{K.P.}},
\bauthor{\bsnm{Chung}, \binits{Y.}},
\bauthor{\bsnm{Li}, \binits{S.-C.}},
\bauthor{\bsnm{Graff}, \binits{D.E.}},
\bauthor{\bsnm{Vermeire}, \binits{F.H.}},
\bauthor{\bsnm{Wu}, \binits{H.}},
\bauthor{\bsnm{Green}, \binits{W.H.}},
\bauthor{\bsnm{McGill}, \binits{C.J.}}:
\batitle{Chemprop: a machine learning package for chemical property prediction}.
\bjtitle{Journal of Chemical Information and Modeling}
\bvolume{64}(\bissue{1}),
\bfpage{9}--\blpage{17}
(\byear{2023})
\end{barticle}
\endbibitem

%%% 30
\bibitem[\protect\citeauthoryear{Breiman}{2001}]{breiman2001random}
\begin{barticle}
\bauthor{\bsnm{Breiman}, \binits{L.}}:
\batitle{Random forests}.
\bjtitle{Machine learning}
\bvolume{45},
\bfpage{5}--\blpage{32}
(\byear{2001})
\end{barticle}
\endbibitem

%%% 31
\bibitem[\protect\citeauthoryear{Gardner and Dorling}{1998}]{gardner1998artificial}
\begin{barticle}
\bauthor{\bsnm{Gardner}, \binits{M.W.}},
\bauthor{\bsnm{Dorling}, \binits{S.}}:
\batitle{Artificial neural networks (the multilayer perceptron)—a review of applications in the atmospheric sciences}.
\bjtitle{Atmospheric environment}
\bvolume{32}(\bissue{14-15}),
\bfpage{2627}--\blpage{2636}
(\byear{1998})
\end{barticle}
\endbibitem

%%% 32
\bibitem[\protect\citeauthoryear{Ramsundar et~al.}{2019}]{Ramsundar-et-al-2019}
\begin{bbook}
\bauthor{\bsnm{Ramsundar}, \binits{B.}},
\bauthor{\bsnm{Eastman}, \binits{P.}},
\bauthor{\bsnm{Walters}, \binits{P.}},
\bauthor{\bsnm{Pande}, \binits{V.}},
\bauthor{\bsnm{Leswing}, \binits{K.}},
\bauthor{\bsnm{Wu}, \binits{Z.}}:
\bbtitle{Deep Learning for the Life Sciences}.
\bpublisher{O'Reilly Media}, \blocation{???}
(\byear{2019})
\end{bbook}
\endbibitem

%%% 33
\bibitem[\protect\citeauthoryear{Weininger}{1988}]{weininger1988smiles}
\begin{barticle}
\bauthor{\bsnm{Weininger}, \binits{D.}}:
\batitle{Smiles, a chemical language and information system. 1. introduction to methodology and encoding rules}.
\bjtitle{Journal of chemical information and computer sciences}
\bvolume{28}(\bissue{1}),
\bfpage{31}--\blpage{36}
(\byear{1988})
\end{barticle}
\endbibitem

%%% 34
\bibitem[\protect\citeauthoryear{}{}]{rdkit}
\begin{botherref}
RDKit: Open-source cheminformatics.
https://www.rdkit.org
\end{botherref}
\endbibitem

%%% 35
\bibitem[\protect\citeauthoryear{Damewood et~al.}{2023}]{damewood2023representations}
\begin{barticle}
\bauthor{\bsnm{Damewood}, \binits{J.}},
\bauthor{\bsnm{Karaguesian}, \binits{J.}},
\bauthor{\bsnm{Lunger}, \binits{J.R.}},
\bauthor{\bsnm{Tan}, \binits{A.R.}},
\bauthor{\bsnm{Xie}, \binits{M.}},
\bauthor{\bsnm{Peng}, \binits{J.}},
\bauthor{\bsnm{G{\'o}mez-Bombarelli}, \binits{R.}}:
\batitle{Representations of materials for machine learning}.
\bjtitle{Annual Review of Materials Research}
\bvolume{53}(\bissue{1}),
\bfpage{399}--\blpage{426}
(\byear{2023})
\end{barticle}
\endbibitem

%%% 36
\bibitem[\protect\citeauthoryear{Ramsundar}{2018}]{ramsundar2018molecular}
\begin{botherref}
\oauthor{\bsnm{Ramsundar}, \binits{B.}}:
Molecular machine learning with deepchem.
PhD thesis,
Stanford University
(2018)
\end{botherref}
\endbibitem

%%% 37
\bibitem[\protect\citeauthoryear{Oliynyk et~al.}{2016}]{oliynyk2016high}
\begin{barticle}
\bauthor{\bsnm{Oliynyk}, \binits{A.O.}},
\bauthor{\bsnm{Antono}, \binits{E.}},
\bauthor{\bsnm{Sparks}, \binits{T.D.}},
\bauthor{\bsnm{Ghadbeigi}, \binits{L.}},
\bauthor{\bsnm{Gaultois}, \binits{M.W.}},
\bauthor{\bsnm{Meredig}, \binits{B.}},
\bauthor{\bsnm{Mar}, \binits{A.}}:
\batitle{High-throughput machine-learning-driven synthesis of full-heusler compounds}.
\bjtitle{Chemistry of Materials}
\bvolume{28}(\bissue{20}),
\bfpage{7324}--\blpage{7331}
(\byear{2016})
\end{barticle}
\endbibitem

%%% 38
\bibitem[\protect\citeauthoryear{Ward et~al.}{2018}]{ward2018matminer}
\begin{barticle}
\bauthor{\bsnm{Ward}, \binits{L.}},
\bauthor{\bsnm{Dunn}, \binits{A.}},
\bauthor{\bsnm{Faghaninia}, \binits{A.}},
\bauthor{\bsnm{Zimmermann}, \binits{N.E.}},
\bauthor{\bsnm{Bajaj}, \binits{S.}},
\bauthor{\bsnm{Wang}, \binits{Q.}},
\bauthor{\bsnm{Montoya}, \binits{J.}},
\bauthor{\bsnm{Chen}, \binits{J.}},
\bauthor{\bsnm{Bystrom}, \binits{K.}},
\bauthor{\bsnm{Dylla}, \binits{M.}}, \betal:
\batitle{Matminer: An open source toolkit for materials data mining}.
\bjtitle{Computational Materials Science}
\bvolume{152},
\bfpage{60}--\blpage{69}
(\byear{2018})
\end{barticle}
\endbibitem

%%% 39
\bibitem[\protect\citeauthoryear{Kraskov et~al.}{2004}]{kraskov2004estimating}
\begin{barticle}
\bauthor{\bsnm{Kraskov}, \binits{A.}},
\bauthor{\bsnm{St{\"o}gbauer}, \binits{H.}},
\bauthor{\bsnm{Grassberger}, \binits{P.}}:
\batitle{Estimating mutual information}.
\bjtitle{Physical Review E—Statistical, Nonlinear, and Soft Matter Physics}
\bvolume{69}(\bissue{6}),
\bfpage{066138}
(\byear{2004})
\end{barticle}
\endbibitem

%%% 40
\bibitem[\protect\citeauthoryear{Shimodaira}{2000}]{shimodaira2000improving}
\begin{barticle}
\bauthor{\bsnm{Shimodaira}, \binits{H.}}:
\batitle{Improving predictive inference under covariate shift by weighting the log-likelihood function}.
\bjtitle{Journal of statistical planning and inference}
\bvolume{90}(\bissue{2}),
\bfpage{227}--\blpage{244}
(\byear{2000})
\end{barticle}
\endbibitem

%%% 41
\bibitem[\protect\citeauthoryear{Shimodaira et~al.}{2009}]{DatasetShiftBook}
\begin{bbook}
\bauthor{\bsnm{Shimodaira}, \binits{H.}},
\bauthor{\bsnm{Sugiyama}, \binits{M.}},
\bauthor{\bsnm{Storkey}, \binits{A.}},
\bauthor{\bsnm{Gretton}, \binits{A.}},
\bauthor{\bsnm{David}, \binits{S.-B.}},
\bauthor{\bsnm{QuinoneroCandela}, \binits{J.}},
\bauthor{\bsnm{Sugiyama}, \binits{M.}},
\bauthor{\bsnm{Schwaighofer}, \binits{A.}},
\bauthor{\bsnm{Lawrence}, \binits{N.}}:
\bbtitle{Dataset Shift in Machine Learning}.
\bsertitle{Neural Information Processing Series},
pp. \bfpage{201}--\blpage{205}.
\bpublisher{Yale University Press in association with the Museum of London}, \blocation{???}
(\byear{2009})
\end{bbook}
\endbibitem

\end{thebibliography}

\end{document}